\documentclass[pdflatex,iicol, sn-mathphys-num]{sn-jnl}

\usepackage{graphicx} % Required for inserting images
\usepackage{booktabs}
\usepackage{tabularx}
\usepackage{makecell}
\usepackage{bbding}
\usepackage{romannum}
\usepackage{pifont}
\usepackage{color,colortbl}
\usepackage{amsmath}
\usepackage{amssymb}
\usepackage{subcaption}
\usepackage{hyperref}
\usepackage{layouts}
\usepackage{microtype}
\usepackage{algorithm}
\usepackage{algorithmic}

\usepackage[table,xcdraw]{xcolor}

\usepackage{pdflscape}   % 横向页面
\usepackage{booktabs}
\usepackage{multirow}
\usepackage{array}
\usepackage{amssymb}     % \checkmark \times
\usepackage{amsmath}
\usepackage{geometry}
\begin{document}

\title[Article Title]{SP-MoMamba: Superpixel-driven Mixture of State Space Experts for Efficient Image Super-Resolution \vspace{1em}}

%\text{\\}

\author[1,2]{\sur{Wenbin Zou}}

\author[1]{\sur{Yawen Cui}}

\author[1]{\sur{Yi Wang}}

\author[1]{\sur{Lap-Pui Chau}}

\author[3]{\sur{Liang Chen}}

\author[4]{\sur{Jinshan Pan}}

\author*[2]{\sur{Huiping Zhuang}}\email{hpzhuang@scut.edu.cn}

\author[5]{Guanbin Li}

%%\affil[1]{\orgdiv{Department of Electrical and Electronic Engineering}, \orgname{The Hong Kong Polytechnic University}, \orgaddress{\city{Hong Kong SAR}}}
\affil[1]{\orgdiv{Department of EEE}, \orgname{The Hong Kong Polytechnic University}, \orgaddress{\city{Hong Kong SAR}}}

\affil[2]{\orgdiv{Shien-Ming Wu School of Intelligent Engineering}, \orgname{South China University of Technology}, \orgaddress{\city{Guangzhou}, \country{China}}}

%%\affil[3]{\orgdiv{School of Electronic Information and Communications}, \orgname{Huazhong University of Science and Technology}, \orgaddress{\city{Wuhan}, \country{China}}}
\affil[3]{\orgdiv{College of Photonic and Electronic Engineering}, \orgname{Fujian Normal University}, \orgaddress{\city{Fuzhou}, \country{China}}}

\affil[4]{\orgdiv{School of Computer Science and Engineering}, \orgname{Nanjing University of Science and Technology}, \orgaddress{\city{Nanjing}, \country{China}}}
\affil[5]{\orgdiv{School of Computer Science and Engineering}, \orgname{Sun Yat-sen University}, \orgaddress{\city{Guangzhou}, \country{China}}}

\date{November 2025}

%\setcounter{page}{0}
%\raggedbottom
\pagenumbering{arabic}

%\begin{center} {\LARGE \textbf{Building Egocentric Procedural AI Assistant: Methods, Benchmarks, and Challenges}}\\[1em] {\large Junlong Li\textsuperscript{1}, Huaiyuan Xu\textsuperscript{1}, Sijie Cheng\textsuperscript{2}, Kejun Wu\textsuperscript{3}, Kim-Hui Yap\textsuperscript{4}, Lap-Pui Chau\textsuperscript{1}, Yi Wang\textsuperscript{1,*}\\[0.5em] } {\small \textsuperscript{1}Department of EEE, The Hong Kong Polytechnic University, Hong Kong SAR\\ \textsuperscript{2}RayNeo, Shenzhen; Tsinghua University, Beijing, China\\ \textsuperscript{3}School of EIC, Huazhong University of Science and Technology, Wuhan, China\\ \textsuperscript{4}School of EEE, Nanyang Technological University, Singapore\\ \textsuperscript{*}Email: yi-eie.wang@polyu.edu.hk }[1em] \end{center} 

\abstract{State space models (SSMs) have emerged as an efficient paradigm for single-image super-resolution (SR) due to their linear complexity and long-range modeling capabilities. However, existing visual SSMs mainly focus on improving how densely represented image features are traversed, while the construction of the visual sequence itself remains largely tied to predefined spatial layouts. Inspired by Gestalt perceptual grouping, we propose SP-MoMamba, a superpixel-driven mixture of state space experts for efficient SR. Instead of performing state-space modeling over densely serialized pixel features, the proposed Superpixel-driven State Space Model (SP-SSM) organizes spatially coherent features into compact region-level tokens and performs global sequence modeling over these content-aware representations, reducing redundant computation while facilitating long-range structural interaction. To accommodate image structures with varying representation granularities, we further develop a Multi-Scale Superpixel Mixture of State Space Experts (MSS-MoE), where scale-specific SP-SSM experts model region-level representations at different granularities and a sparse router dynamically selects an appropriate modeling scale. In addition, a Local Spatial Modulation Expert (LSME) complements region-level global modeling by refining local high-frequency details. Extensive experiments demonstrate that SP-MoMamba achieves strong reconstruction performance with a favorable trade-off among model size, computational cost, and inference efficiency.
}


\maketitle
% \footnote{This work was completed under the joint guidance of Huiping Zhuang and Yi Wang.}

\section{Introduction}
% Single image super-resolution (SR) is a fundamental technique in image processing, designed to reconstruct high-resolution (HR) images from low-resolution (LR) counterparts, thereby enhancing image detail and visual quality. This technology is widely applied across diverse fields, including medical imaging, surveillance systems, and satellite imagery. Numerous studies have leveraged convolutional neural networks (CNNs) \cite{SRCNN, EDSR, RCAN} and Transformers \cite{swinir, GRL, srformer} to learn this inherently ill-posed mapping relationship. However, the majority of SR methods \cite{EDSR, RCAN} rely on deep and complex architectures to achieve superior performance. These approaches often entail high computational complexity, rendering real-time processing impractical on resource-constrained devices, which limits their deployment and widespread adoption in real-world scenarios. Although various strategies, including neural architecture search \cite{nas}, recursive networks \cite{DRRN}, and model distillation \cite{IDN, IMDN}, have been explored to improve efficiency, achieving a favorable balance between reconstruction quality and computational cost remains challenging.
Single-image super-resolution (SR) aims to reconstruct high-resolution (HR) images from their low-resolution (LR) counterparts and has been widely applied in medical imaging, surveillance, and remote sensing. Numerous studies have leveraged convolutional neural networks (CNNs) \cite{SRCNN, EDSR, RCAN} and Transformers \cite{swinir, GRL, srformer} to learn this inherently ill-posed mapping. Despite remarkable progress, many high-performance SR methods \cite{EDSR, RCAN} rely on increasingly sophisticated architectures, resulting in considerable computational and memory overhead. This limits their practical deployment, particularly on resource-constrained devices. Although various strategies, including neural architecture search \cite{nas}, recursive networks \cite{DRRN}, and model distillation \cite{IDN, IMDN}, have been explored to improve efficiency, achieving a favorable balance between reconstruction quality and computational cost remains challenging.

\begin{figure*}[t]
  \centering
  \includegraphics[width=\textwidth]{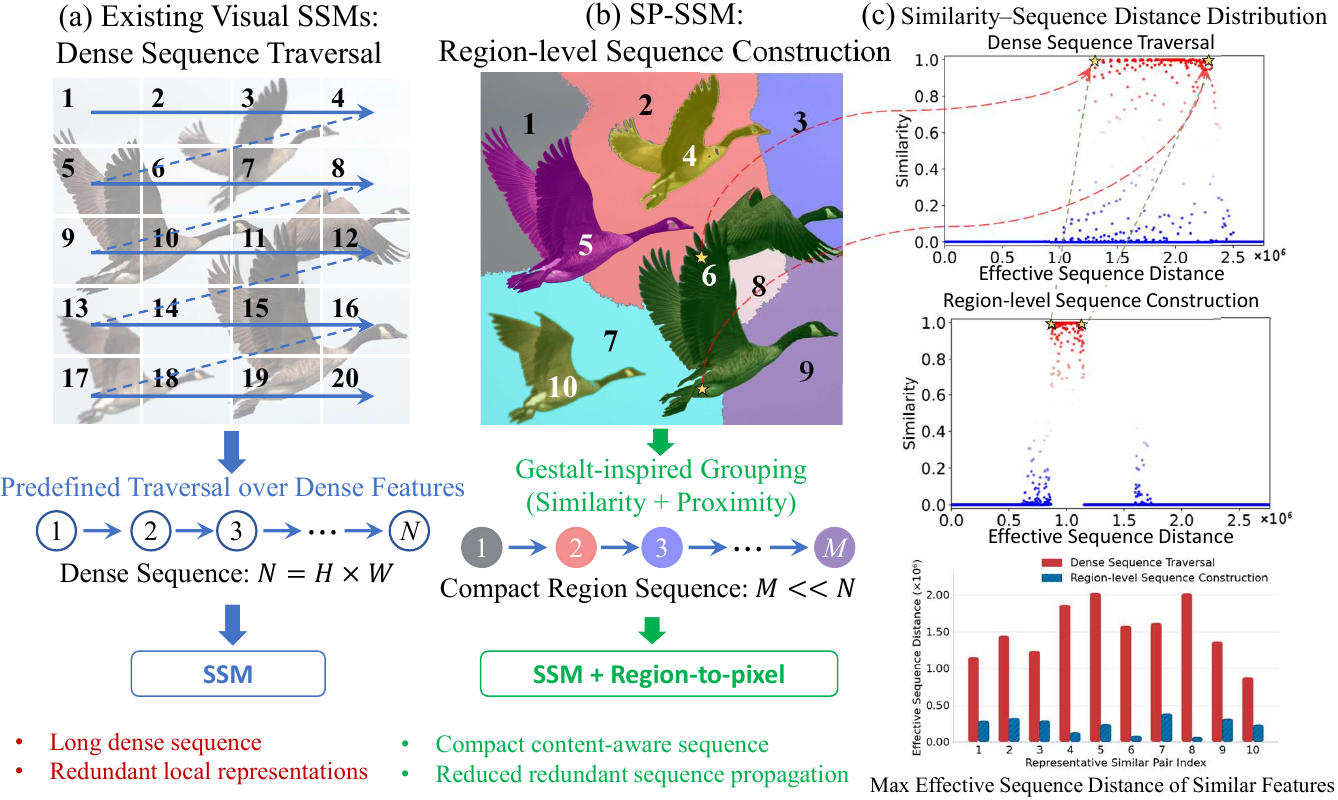}
  \caption{Motivation of SP-SSM. (a) Conventional visual SSMs perform predefined traversal over dense spatial features, producing long sequences with redundant local representations. (b) SP-SSM organizes spatially coherent features into compact region-level representations through Gestalt-inspired grouping before state-space modeling. (c) Similarity--sequence-distance statistics show that region-level construction substantially shortens the effective sequence distance among highly similar features compared with dense traversal, providing statistical motivation for content-aware visual sequence construction.}
  % (c) Comparison between performance vs Inference times and GPU Memory on Manga109 ×4 dataset. Inference times and GPU Memory are calculated on 720p HR image.
  \label{fig:motivation}
  \vspace{-5pt}
\end{figure*}

Recently, State Space Models (SSMs), represented by Mamba \cite{mamba}, have emerged as a promising alternative for efficient image restoration due to their linear sequence complexity and strong long-range modeling capability. Following their success in high-level vision tasks such as image classification \cite{vmamba, visonmamba} and object detection \cite{obmamba, mambayolo}, visual SSMs have been increasingly introduced into low-level vision tasks, including image denoising \cite{mambair}, SR \cite{hi-mamba}, and low-light image enhancement \cite{wave-mamba, fremamba}. For example, MambaIR \cite{mambair} demonstrates that state-space modeling can achieve competitive restoration performance with favorable computational efficiency. These advances highlight the potential of SSMs for efficient SR, while also raising an important question: \textbf{how should visual information be represented and organized for effective sequential state-space modeling?}

Despite these advances, many existing Mamba-based SR methods retain a common modeling paradigm: state-space modeling is performed over densely represented image features, while most architectural efforts focus on improving how these features are traversed. Multi-directional scanning \cite{mambair}, cross-scanning \cite{fremamba}, and other enhanced traversal strategies effectively alleviate the directional limitations of individual scan paths and improve spatial information propagation. However, they generally preserve the underlying dense spatial representation and primarily redesign the order in which spatial features participate in sequential modeling. As illustrated in Fig.~\ref{fig:motivation}(a), densely represented spatial features are serialized individually, producing long visual sequences in which many neighboring elements exhibit highly correlated appearance and structural information. Consequently, considerable redundant information may still be repeatedly propagated during state-space modeling. More importantly, the resulting sequence organization is predominantly determined by predefined spatial layouts and traversal rules rather than by the content-dependent organization of the input image.

To further examine this issue independently of any trained network, we conduct a preliminary image-statistical analysis in Fig.~\ref{fig:motivation}(c). Specifically, we compare predefined dense traversal with content-adaptive region construction by measuring the similarity--sequence-distance distribution of visually similar elements. Under dense traversal, highly similar features can remain widely separated in the serialized sequence. In contrast, superpixel grouping aggregates spatially proximate and appearance-consistent elements into shared region representations, substantially reducing their effective sequence distance. The same tendency is observed for representative high-similarity pairs, whose effective sequence distances are consistently shortened after region-level construction. This analysis suggests that the limitation lies not only in how dense features are traversed, but also in what representation is presented to the SSM. This observation motivates a more fundamental question: \textbf{rather than continually redesigning traversal paths over dense image features, can the visual sequence itself be constructed according to image content before state-space modeling?}

To explore this direction, we draw inspiration from Gestalt perceptual grouping theory \cite{Gestalt1}, which suggests that spatially proximate visual elements with similar appearance tend to be organized into coherent perceptual units. This perceptual organization provides a natural motivation for constructing visual sequences according to image content rather than predefined spatial traversal. Based on this perspective, we propose SP-MoMamba, a superpixel-driven mixture of state space experts for efficient SR. Importantly, Gestalt theory motivates how visual elements are organized before sequential modeling, rather than prescribing the architecture of the state-space model itself.

Following this principle, we introduce the Superpixel-driven State Space Model (SP-SSM) as the core component of SP-MoMamba. As illustrated in Fig.~\ref{fig:motivation}(b), instead of directly performing state-space modeling over densely represented spatial features, SP-SSM first organizes spatially coherent features into a compact set of region-level representations and then performs global sequence modeling over these content-aware regions. Superpixel grouping therefore serves as a differentiable computational realization of Gestalt-inspired perceptual organization, where spatial proximity and appearance similarity jointly determine the region construction. By transforming the original dense sequence of $N=H\times W$ spatial elements into a compact region sequence of $M$ elements ($M\ll N$), SP-SSM shortens the effective sequence distance among locally similar features and reduces redundant sequential propagation before state-space modeling. The enhanced region representations are subsequently projected back to the dense spatial domain for reconstruction.

Region-level sequence modeling, however, introduces an additional design factor: the appropriate representation granularity varies considerably across image structures. Large homogeneous regions can be represented efficiently at a coarse granularity, whereas fine and repetitive textures require more detailed region representations. A single fixed grouping scale may therefore be insufficient for heterogeneous image content. To address this granularity mismatch, we develop the Multi-Scale Superpixel Mixture of State Space Experts (MSS-MoE). MSS-MoE constructs multiple scale-specific SP-SSM experts, each corresponding to a distinct granularity of region-level sequence modeling. A sparse router dynamically selects the appropriate expert according to the input features, thereby adapting the granularity at which global state-space modeling is performed while avoiding unnecessary activation of all experts.

While compact region-level abstraction benefits efficient global modeling, aggregating neighboring features into region representations may attenuate fine spatial variations that are critical for SR reconstruction. We therefore introduce the Local Spatial Modulation Expert (LSME) as a complementary local refinement pathway. LSME focuses on fine spatial dependencies and high-frequency structures in the dense feature domain, complementing the region-level global interactions modeled by SP-SSM and MSS-MoE. Together, these components establish a complementary region-level global and dense local modeling framework for efficient and detailed image reconstruction.

Our contributions are summarized as follows:

\begin{itemize}
    \item We introduce SP-SSM, a content-aware region-level state-space modeling approach that reformulates how visual sequences are constructed before sequential modeling. Instead of exhaustively modeling densely represented spatial features, SP-SSM organizes spatially coherent features into compact region-level representations, shortening the effective sequence distance among locally similar features and reducing redundant sequential modeling.

    \item We develop MSS-MoE to address the granularity mismatch inherent in region-level modeling. Scale-specific SP-SSM experts operate at different representation granularities, while sparse routing dynamically selects the appropriate modeling scale according to the input features, enabling adaptive multi-granularity global modeling with efficient expert activation.

    \item Extensive experiments on synthetic and real-world SR benchmarks demonstrate the effectiveness of the proposed region-level modeling framework, achieving strong reconstruction performance with a favorable balance among model size, computational cost, and inference efficiency. Further experiments on multiple image restoration tasks validate its generalization capability.
\end{itemize}

\section{Related Works}\label{sec:relatedwork}
\subsection{Deep Networks for Image Super-Resolution}

Following the introduction of SRCNN \cite{SRCNN}, which pioneered the application of convolutional neural networks to the field of image SR and achieved remarkable success, a growing number of researchers \cite{ESPCNN, EDSR, RCAN, jwsgn} have explored the integration of advanced deep learning architectures into this domain. For example, VDSR \cite{VDSR} and EDSR \cite{EDSR} employed residual learning to deepen network architectures, significantly enhancing the quality of image restoration. Subsequently, leveraging the exceptional performance of Transformers in high-level tasks, SwinIR \cite{swinir} incorporated the local window self-attention and shift mechanisms of the Swin Transformer \cite{swintrans} into image restoration, further improving the quality of reconstructed images. Inspired by the success of SwinIR, researchers have focused on designing sophisticated self-attention mechanisms, such as ART \cite{art}, OmniSR \cite{omnisr}, and GRL \cite{GRL}, to capture global information more effectively. Although these methods achieve impressive metric scores, they often demand substantial computational resources and a large number of parameters, posing challenges to their practical deployment on resource-constrained edge devices.

\subsection{Efficient Super-Resolution Methods}
To improve the practicality of SR models, extensive studies have explored lightweight CNN architectures \cite{carn, IMDN, SAFMN} and efficient Transformers \cite{ELAN, ESRT, scet}. CARN \cite{carn} employs cascading connections to enhance feature reuse, IMDN \cite{IMDN} reduces redundancy through feature distillation, and SAFMN \cite{SAFMN} constructs multi-scale feature modulation with a lightweight architecture. For Transformer-based SR, ELAN \cite{ELAN}, ESRT \cite{ESRT}, and SCET \cite{scet} reduce attention complexity through efficient spatial or channel-wise interaction. More recently, PFT \cite{PFT} progressively filters irrelevant token interactions to reduce unnecessary attention computation, while SAT \cite{SAT} selectively aggregates key-value tokens to enable efficient long-range interaction.

Region-aware representations have also been explored for efficient image restoration. CODE \cite{code} first models global dependencies at the superpixel level and subsequently transfers the aggregated global information back to individual pixels. SPIN \cite{SPIN} employs superpixels to construct locally coherent regions and performs intra-superpixel and superpixel cross-attention for lightweight SR. These studies demonstrate the effectiveness of superpixels for reducing redundant pixel-wise interaction. However, they primarily employ superpixel representations within attention-based architectures. In contrast, our work investigates their role in visual sequence construction for state-space modeling, where content-adaptive regions constitute the compact sequence over which the SSM performs global interaction.

% In the pursuit of more efficient models, researchers have introduced lightweight architectures \cite{carn, IMDN, SAFMN} and efficient Transformers \cite{ELAN, ESRT, scet}. For instance, CARN \cite{carn} proposes a cascading mechanism that integrates local and global features to enhance network efficiency. IMDN \cite{IMDN} employs a feature distillation approach to split and aggregate features, further reducing both parameter count and computational load. SAFMN \cite{SAFMN} improves network efficiency by constructing a feature pyramid based on channel utilization. Meanwhile, to mitigate the high complexity introduced by Transformer architectures, ELAN \cite{ELAN} and ESRT \cite{ESRT} reduce feature dimensionality through segmentation or scaling, thereby improving computational efficiency. And SCET \cite{scet} further minimizes computational costs by designing a pixel attention mechanism and an efficient Transformer module. More recently, SPIN \cite{SPIN} leverages superpixel attention to decrease the computational burden of Transformers while employing cross-attention to simultaneously capture local and global information. Although these methods are lightweight and efficient, there remains room for improvement in balancing the trade-off between efficiency and performance.

\subsection{Mamba for Image Restoration}
State Space Models (SSMs) have recently emerged as an efficient alternative for visual long-range modeling due to their linear sequence complexity. Vision-oriented architectures such as VMamba \cite{vmamba} and VIM \cite{visonmamba} adapt selective state-space modeling to two-dimensional visual representations, and subsequent works have extended this paradigm to various image restoration tasks \cite{fremamba, wave-mamba}. MambaIR \cite{mambair}, for example, combines visual state-space modeling with channel interaction to achieve competitive restoration performance with favorable computational efficiency.

A major line of research has focused on improving how visual features are organized and traversed by SSMs. MambaIRv2 \cite{mambairv2} introduces Attention-guided State-space Equations and Semantic-Guided Neighborhoods to alleviate the causal constraints associated with sequential scanning. EVSSM \cite{EVSSM} employs geometric transformations to enhance non-local information interaction through modified traversal patterns. These approaches improve the propagation of spatial information while largely operating on densely represented visual features.

More recently, C$^2$SSM \cite{C2SSM} takes a related but distinct direction by replacing pixel-serial scanning with cluster-centric modeling, where dense features are summarized into a sparse set of cluster centroids for state-space interaction and the resulting global context is subsequently propagated back to pixels. In contrast to centroid-based feature clustering, SP-SSM constructs the SSM sequence through differentiable superpixel grouping, explicitly organizing spatially coherent features into content-adaptive region representations. Furthermore, MSS-MoE extends this formulation by dynamically selecting among scale-specific SP-SSM experts, allowing the granularity of region-level state-space modeling to adapt to the input features.

\begin{figure*}[t]
    \centering
    \includegraphics[width=0.7\textwidth]{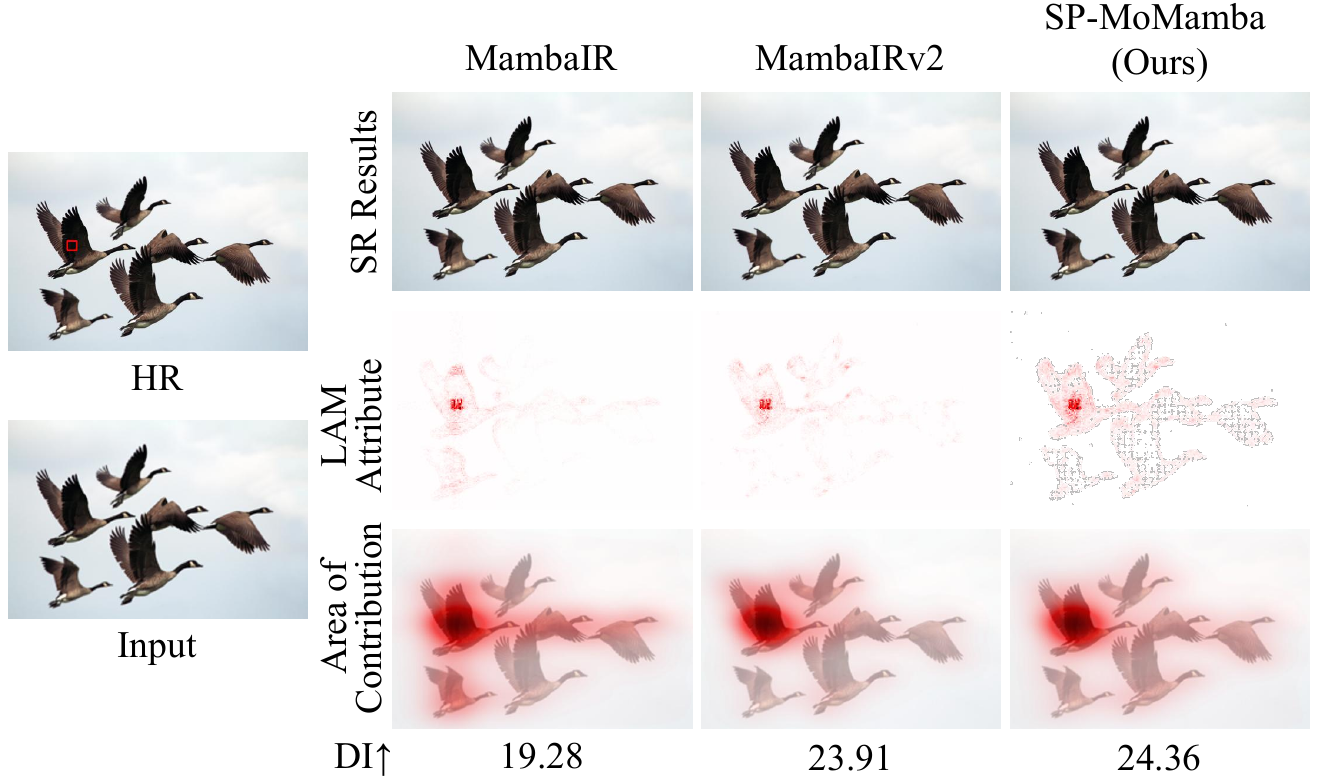}
    \caption{Comparison of long-range information utilization using Local Attribution Maps (LAM) and the Diffusion Index (DI). A higher DI indicates a broader spatial range of features contributing to the reconstruction of the selected target region.}
    \label{fig:asp}
\end{figure*}

\subsection{Mixture of Experts for Image Restoration}
Mixture-of-Experts (MoE) architectures provide a flexible mechanism for conditional computation and have been widely explored in both high-level vision \cite{moeob, moeclass} and image restoration \cite{moesr, seemore,swin2mose}. Existing restoration-oriented MoE methods typically specialize experts according to degradation characteristics, feature subspaces, or computational capacity. For example, degradation-aware methods \cite{moesr, moe57} route features according to latent degradation representations, while SeemoRe \cite{seemore} employs rank-modulated experts together with spatial modulation for efficient SR. Swin2-MoSE \cite{swin2mose} integrates expert routing into a Transformer-based restoration architecture.

More recently, MoCE-IR \cite{MOCEIR} introduces complexity experts with different computational costs and receptive fields and learns task-discriminative routing to allocate restoration tasks to experts with appropriate capacity. While these methods demonstrate the effectiveness of conditional expert selection, their expert specialization is mainly associated with degradation/task characteristics or network complexity. In contrast, the experts in our MSS-MoE share the same SP-SSM formulation but operate at different region-level representation granularities. The router therefore determines the granularity at which global state-space modeling is performed, enabling content-dependent adaptation of the visual sequence representation while activating only a sparse subset of experts.

% The Mixture of Experts (MoE) paradigm has gained widespread adoption in large-scale language models due to its efficiency and scalability. Consequently, MoE has been extended to advanced vision tasks, including image classification \cite{moeob} and object detection \cite{moeclass}, as well as low-level vision tasks \cite{moesr, seemore,swin2mose}. For instance, studies \cite{moesr} and \cite{moe57} extract latent degradation features to construct MoE-based adaptive networks, effectively addressing diverse degradation patterns in blind SR. SeemoRe \cite{seemore} employs rank-modulated experts to prioritize features with the highest information content, followed by spatial modulation experts to achieve precise spatial enhancement. Similarly, Swin2-MoSE \cite{swin2mose} enhances Swin2SR \cite{swin2sr} by incorporating an MoE framework, yielding improved visual outcomes. While these methods leverage the flexibility and efficiency of MoE to achieve commendable performance, there remains potential for further improvement in the quality of SR reconstruction.
\section{Motivation}\label{motivation}
\subsection{State Space Models}
% State Space Models (SSMs) \cite{mamba} are mathematical models used in control theory and signal processing to describe dynamic systems. The following equations define standard SSMs:
State Space Models (SSMs) \cite{mamba} describe the evolution of a latent state driven by an input sequence. A continuous SSM can be formulated as:
\begin{align}
    h'(t) &= \mathbf{A}h(t)+\mathbf{B}x(t), \\ 
    y(t) &= \mathbf{C}h(t) + \mathbf{D}x(t),
\end{align}
where $\mathbf{A}, \mathbf{B}, \mathbf{C},$ and $\mathbf{D}$ represent the system parameters. The terms $x(t), h(t)$, and $y(t)$ denote the input, hidden state, and output, respectively. In the context of deep learning, existing SSM-based methods, such as Mamba, employ a zero-order hold (ZOH) to discretize continuous state space equations. This discretization enables the efficient processing of long sequence data and is applicable to various sequence modeling tasks. The discrete representation is governed by the following equations:
\begin{align} \label{eq}
    h_t &= \bar{\mathbf{A}}h_{t-1}+\bar{\mathbf{B}}x_t, \\ 
    y_t &= \mathbf{C}h_t + \mathbf{D}x_t, \\
    \bar{\mathbf{A}} &= \text{exp}(\Delta \mathbf{A}), \\
    \bar{\mathbf{B}} &= (\Delta \mathbf{A})^{-1}(\text{exp}(\Delta \mathbf{A}) - \mathbf{I})(\Delta \mathbf{B}),
\end{align}
where $\bar{\mathbf{A}}$ and $\bar{\mathbf{B}}$ are the discrete counterparts of $\mathbf{A}$ and $\mathbf{B}$. The symbol $\Delta$ denotes the discretization timescale parameter, which is utilized to transform the continuous parameters $\mathbf{A}$ and $\mathbf{B}$ into their discrete forms $\bar{\mathbf{A}}$ and $\bar{\mathbf{B}}$.

To enhance dynamic feature modeling, Mamba \cite{mamba} adopts an input-aware parameterization for $\mathbf{B}$, $\mathbf{C}$, and $\Delta$. Similar to the recursive formulation in Eq. (\ref{eq}), Mamba excels at propagating information through exceptionally long sequences, a capability that is pivotal for integrating global pixel information during restoration. However, unlike standard recurrent architectures, Mamba utilizes a hardware-aware parallel scan algorithm \cite{mamba}. This approach inherits the parallel processing advantages of Eq. (\ref{eq}), significantly optimizing the efficiency of both training and inference.

\subsection{From Dense Visual Sequences to Content-Aware Region Modeling}
% To extend one-dimensional SSMs to two-dimensional visual tasks, the vast majority of existing methods \cite{mambair, mambairv2, EVSSM, hi-mamba}, rely on the two-dimensional selective scanning mechanism (SS2D) \cite{vmamba} or its multi-directional variants. From an information-theoretic perspective, however, this mechanism suffers from a fundamental limitation. Natural images exhibit extremely high spatial redundancy at the pixel level, yet existing grid-based scanning forcibly flattens continuous two-dimensional visual signals into fixed-length one-dimensional sequences. This data-agnostic rigid scanning inevitably severs the inherent topological structure of objects. As illustrated in Fig. \ref{fig:enter-label}(a), physically adjacent regions are stretched to extreme distances after forced serialization, causing long-range dependency decay within excessively long and redundant sequences. Simultaneously, fixed-grid partitioning fragments originally coherent semantic regions, such as repetitive sky or water textures, into disconnected pieces, which readily induces semantic confusion.
To adapt one-dimensional SSMs to two-dimensional visual representations, existing methods \cite{mambair, mambairv2, EVSSM, hi-mamba} have developed selective, multi-directional, and other enhanced scanning strategies to facilitate spatial information propagation. While these strategies improve how spatial features are traversed, they generally retain a dense visual representation in which individual spatial features participate separately in state-space modeling. However, natural images contain substantial local redundancy: spatially neighboring features frequently exhibit similar appearance and structural characteristics. Consequently, dense visual sequences repeatedly propagate highly correlated information and remain largely organized according to predefined spatial layouts rather than image content. Although SSMs exhibit linear complexity, their computational cost and information propagation still scale with sequence length. Therefore, representing highly correlated spatial locations individually can introduce unnecessary sequential computation and make global modeling operate over a substantially longer sequence than necessary. More importantly, a dense spatial sequence does not explicitly encode which locations belong to the same coherent image structure. This motivates a representation in which structurally consistent features can first be organized into compact regions before global sequential modeling.

% To bridge this gap, we resort to the perceptual grouping theory in Gestalt psychology \cite{Gestalt1}. The human visual system naturally tends to aggregate pixels sharing similar color and texture into coherent entities. Motivated by this theory, we propose using superpixels as the input primitives of the SSM. Unlike naive grid-based patch partitioning, superpixel clustering is essentially a structure-aware information bottleneck: it filters out low-level local redundancy while maximally preserving the core semantic manifold of the image. Through this process, the low-level dense pixel sequence is elevated into a high-level, compact semantic token sequence. This not only substantially reduces the sequence length of the SSM, but also liberates it entirely from meaningless spatial exhaustion, enabling the model to focus on semantic interactions among global visual entities in a manner analogous to processing natural language tokens.

This perspective is closely related to perceptual grouping in Gestalt theory \cite{Gestalt1}, where spatial proximity and visual similarity encourage individual elements to be perceived as coherent units. We use this principle as a representation-level motivation for organizing visual features prior to state-space modeling. Following this motivation, we employ differentiable superpixel grouping as a computational mechanism to aggregate spatially proximate and visually similar features into content-adaptive region representations. Unlike regular grid partitioning, the resulting regions adapt their spatial support according to the input features while preserving local spatial coherence. Through this grouping process, the original dense representation containing $N=H\times W$ spatial elements is converted into a compact sequence of $M$ region-level representations, where $M<<N$. State-space modeling is subsequently performed over these compact region representations rather than the original dense spatial sequence. This formulation substantially shortens the effective sequence length and allows long-range interactions to be modeled among content-aware regions, while the learned pixel-to-region associations preserve a pathway for projecting the enhanced representations back to the dense spatial domain.

However, region-level representation introduces an additional design factor that is absent from dense pixel-wise modeling: grouping granularity. Different image structures require different levels of abstraction. Large homogeneous areas can be represented efficiently with coarse grouping, whereas intricate and repetitive textures benefit from finer region representations. Therefore, a fixed grouping scale may not provide an optimal representation for heterogeneous image content. This observation motivates the adaptive multi-granularity modeling strategy introduced in our MSS-MoE.

% To extend SSMs from 1D sequence data to 2D visual data, the majority of current research \cite{mambair, hi-mamba, vmamba} employs a 2D selective scanning mechanism (SS2D) \cite{vmamba} to capture spatial correlations, as illustrated in Fig. \ref{fig:enter-label}(a). However, the process of flattening an image into a 1D sequence frequently disrupts inherent semantic relationships. For example, spatially adjacent objects in Fig. \ref{fig:enter-label}(a) may be widely separated in the resultant 1D sequence, thereby hindering the ability of the model to leverage proximity for semantic inference. Furthermore, images frequently contain repetitive structures—such as skies and buildings—that share similar textures, which heightens the risk of semantic confusion. Once unfolded, information from these structures may be incorrectly associated, leading to erroneous predictions. Consequently, current SS2D methods struggle to adequately preserve critical spatial structure and semantic information.

% In comparison to traditional SS2D, which systematically transforms 2D features into 1D sequences, superpixel sampling clusters semantically similar pixels based on color or texture. This approach effectively reduces the number of pixels requiring processing while maintaining the spatial structure and semantic integrity of the image. Therefore, integrating superpixel algorithms into SSMs provides a robust solution to the limitations of conventional SSMs when processing 2D images. 

\begin{figure*}[t]
    \centering
    \includegraphics[width=0.8\textwidth]{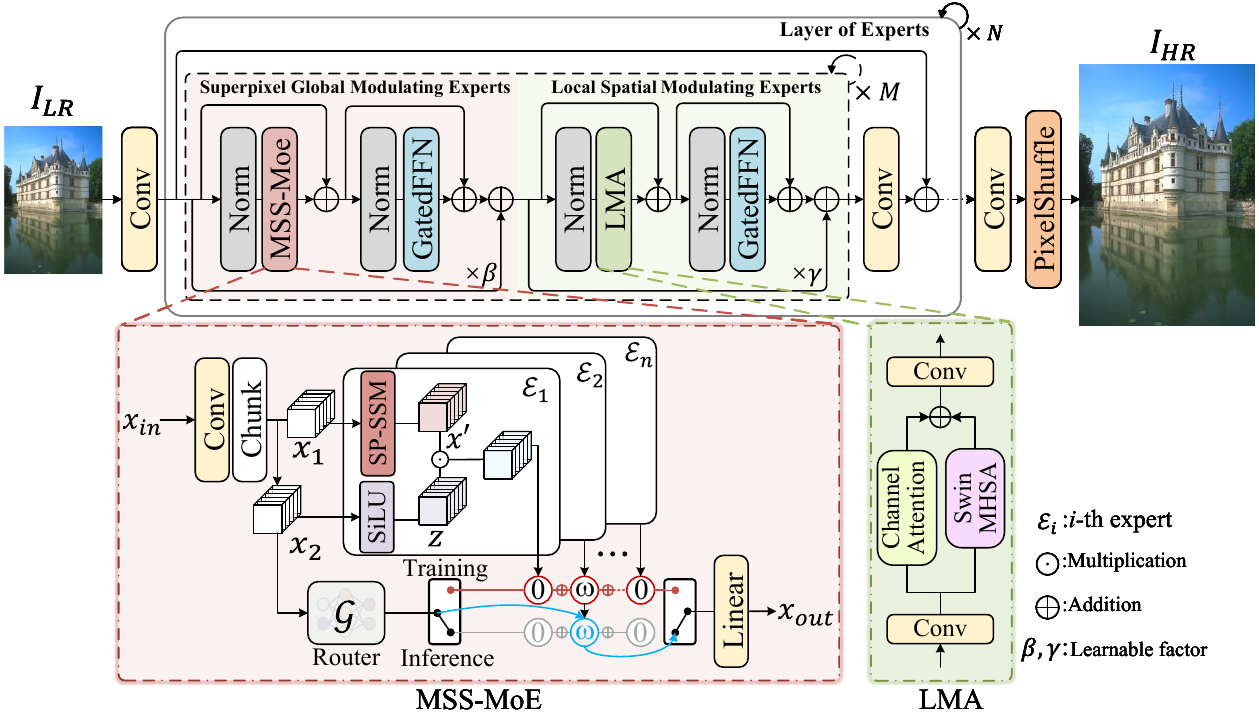}
    \caption{Overall architecture of SP-MoMamba. Each Layer of Experts (LoE) combines adaptive region-level global modeling and complementary local spatial refinement. SP-SSM constructs compact content-aware region sequences for state-space modeling, while MSS-MoE dynamically selects the appropriate representation granularity through scale-specific experts. LSME complements the region-level abstraction by refining local high-frequency structures.}
    \label{fig:network}
\end{figure*}

\subsection{Visual Analysis of Long-Range Structural Aggregation}
To examine how region-level sequence modeling affects long-range information aggregation, we analyze the receptive fields and information utilization patterns of different models using Local Attribution Maps (LAM) and the Diffusion Index (DI) \cite{LAM}, as shown in Fig. \ref{fig:asp}. When reconstructing the selected bird region, MambaIR \cite{mambair} and MambaIRv2 \cite{mambairv2} exhibit relatively localized attribution patterns, indicating that their reconstruction relies more heavily on nearby spatial evidence. In comparison, SP-MoMamba activates a broader set of spatially separated but structurally similar regions. The activated regions of SP-MoMamba extend beyond the local neighborhood and show stronger responses around other visually similar bird structures in the image. This suggests that the proposed region-level modeling facilitates the utilization of spatially distant structures when they provide relevant reconstruction evidence.

For the selected target, the attribution map indicates that SP-MoMamba draws information from several spatially separated regions with similar visual structures, which is consistent with the intended long-range interaction among region-level representations. This broader information utilization is also reflected by the higher DI score of 24.36, compared with 19.28 for MambaIR and 23.91 for MambaIRv2. These observations provide empirical evidence that region-level sequence modeling can broaden the effective information aggregation range and facilitate interactions among structurally related image regions.

\section{Methodology}\label{method}

% In this section, we present our proposed SP-MoMamba, as illustrated in Fig. \ref{fig:network}. The complete architecture of our pipeline integrates $N$ layers of experts (LoEs) and upsampling layers. Initially, a $3\times 3$ convolutional operation is employed to extract shallow features from the input low-resolution image. These features are then processed through a series of LoEs to recover deep features. Each LoE consists of $M$ paired sets of Superpixel Global Modulating Experts (SGME) and Local Spatial Modulating Experts (LSME), collaboratively enhancing feature restoration. SGME adopts a collaborative reconstruction method by integrating a multi-scale superpixels mixture of state space experts (MSS-MoE), maximizing the interaction of global information. LSME concentrates on refining local features through a localized mixed attention mechanism, which enhances overall performance. In addition, two residual connections with learnable scales $\beta,$ and $\gamma$ are introduced. Finally, the refined deep features are transformed into high-resolution images via a pixelshuffle and convolution.

In this section, we introduce the architecture of SP-MoMamba, as illustrated in Fig. \ref{fig:network}. Given a low-resolution input image, a $3\times 3$ convolution is first employed to extract shallow features, which are subsequently refined through $N$ stacked Layers of Experts (LoEs) for deep feature reconstruction. Each LoE consists of two complementary branches: a Superpixel Global Modulating Expert (SGME) for adaptive region-level global modeling and a Local Spatial Modulating Expert (LSME) for local high-frequency refinement. Specifically, SGME is built upon the proposed MSS-MoE, which performs state-space modeling on content-aware regional representations at adaptively selected granularities. In parallel, LSME captures fine-grained spatial dependencies in the dense feature domain, complementing the region-level representations learned by SGME and enhancing local detail recovery. Finally, the globally and locally refined features are fed into a reconstruction module to generate the high-resolution output.

\subsection{Superpixel Global Modulating Experts}
% Unlike prior works \cite{SPIN, mambairv2, EVSSM, C2SSM}, which rely on substantial computational resources, we prioritize efficiency by constructing global similarity relationships based on interactions among the most relevant scale-specific superpixels. As shown in Fig. \ref{fig:network}, we propose the MSS-MoE, which improves multi-scale modeling while keeping the dominant inference cost bounded by the selected top-k experts. MSS-MoE employs superpixel-driven state space models (SP-SSMs) at different scales to independently model global features across different resolutions. By leveraging the strengths of the mixture of experts scheme, it selectively integrates the resulting features, ensuring optimal global modeling within each LoE. Then, a Gated Feed-Forward Network (GatedFFN) \cite{dat} is utilized to aggregate contextual information from these global features.
The Superpixel Global Modulating Expert is designed to perform efficient long-range modeling over compact region-level representations. Rather than directly processing all dense spatial features during global interaction, SGME first constructs region representations through SP-SSM and then adaptively determines the modeling granularity through MSS-MoE. As shown in Fig. \ref{fig:network}, we propose the MSS-MoE, which improves multi-scale modeling while keeping the dominant inference cost bounded by the selected top-k experts. MSS-MoE employs superpixel-driven state-space models (SP-SSMs) at different scales to independently model global features across different resolutions. By leveraging the strengths of the mixture of experts scheme, it selectively integrates the resulting features, ensuring optimal global modeling within each LoE. Then, a Gated Feed-Forward Network (GatedFFN) \cite{dat} is utilized to aggregate contextual information from these global features.

\begin{figure*}[t]
    \centering
    \includegraphics[width=0.9\textwidth]{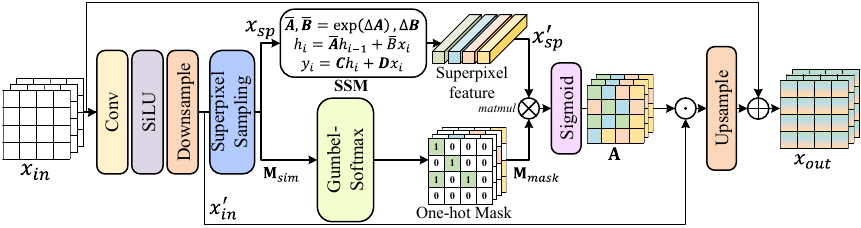}
    \caption{Illustration of the superpixel-driven state-space model (SP-SSM). The one-hot mask has a shape of $N \times M$, which is reshaped into a tensor of shape $H_s \times W_s \times M$ for ease of understanding in the figure. }
    \label{fig:SPSM}
\end{figure*}

\subsubsection{Multi-scale Superpixel Mixture of State Space Experts (MSS-MoE)}
% As shown in Fig. \ref{fig:network}, the output features from the last layer normalization serve as the input features $x_{in}$ for this module. A Linear layer is first applied to increase the dimensionality of the feature channels, followed by a split along the channel dimension to yield two distinct features, $x_1$ and $x_2$. Subsequently, we employ a SP-SSM module to derive the global attention feature $x'$ from $x_1$. Meanwhile, $x_2$ is processed through an activation function to obtain the gating feature $z$. Consequently, the formulation for each superpixel state space expert is expressed as follows: where $\text{SP-SSM}(\cdot)$ and $\sigma(\cdot)$ denote the SP-SSM module and the SiLU \cite{silu} activation function, respectively. The $\odot$ denotes Hadamard product. The $s$ represents the scale parameter of SP-SSM. We employ an SP-SSM to ensure robust modeling of global information while introducing an residual connection to prevent the loss of local information.
% However, the SP-SSM operating at a fixed scale may fail to fully exploit all internal information, thereby limiting the model’s expressive capacity. To address this, we propose an ensemble approach that integrates superpixel state space experts across multiple scales $s_i$. A routing network searches the solution space to identify the optimal scale for the superpixel state space experts based on the input and network depth. The final output $x_{out}$ of the MSS-MoE is formulated as follows:
As shown in Fig. \ref{fig:network}, given the normalized input feature $x_{in}$, a linear projection expands the channel dimension and splits the feature into two branches, $x_1$ and $x_2$. The first branch is processed by a scale-specific SP-SSM for region-level global modeling, while the second branch generates a gating feature through a SiLU activation. For the $i-$th expert,
\begin{equation}
    \mathcal{E}_i(x_1,x_2,s_i)=x' \odot z = \text{SP-SSM}(x_1, s_i)\odot \sigma(x_2),
\end{equation}

where $s_i$ controls the effective granularity of region-level sequence construction in the $i$-th expert.

A fixed SP-SSM scale imposes a single region-level representation granularity, which may be insufficient for image structures with heterogeneous spatial extents and texture complexity. MSS-MoE therefore constructs several SP-SSM experts with different scale parameters $\{s_i\}_{i=1}^n$, corresponding to different region-level modeling granularities. A routing network predicts the relevance of each granularity from the current feature representation and dynamically selects the most suitable experts for computation. The output is formulated as

\begin{equation}
    x_{out} = \sum_i^n \mathcal{G}(x_2)\mathcal{E}_i(x_1, x_2, s_i),
\end{equation}
where $\mathcal{G}(x_2)$ denotes the routing weight assigned to the $i$-th expert. The router consists of a linear projection followed by Softmax. During sparse inference, only the top-$k$ experts with the highest routing scores are activated, enabling adaptive region-level modeling while limiting the dominant expert computation to the selected granularities.

\subsubsection{Superpixel-driven State Space Model (SP-SSM)}
The key distinction of SP-SSM lies in how the visual sequence is represented before state-space modeling. Rather than applying the SSM directly to all densely represented spatial features, SP-SSM first aggregates spatially coherent features into a compact region-level sequence, performs global state-space modeling in this reduced representation space, and subsequently projects the enhanced region information back to the dense spatial domain. 

As shown in Fig. \ref{fig:SPSM}, given the input feature $x_{in} \in \mathbb{R}^{H\times W\times C}$, we use $3 \times3 $ convolution and SiLU activation function to map features. The transformed features are then downsampled by a scale factor $s$ and grouped through differentiable superpixel sampling, producing $M$ region-level features $x_{sp}\in \mathbb{R}^{M\times C}$ together with a pixel-to-region similarity matrix $\textbf{M}_{sim} \in \mathbb{R}^{N_s\times M}$, where $N_s=H_s\times W_s, H_s=H/s, W_s=W/s$:
% Then, these features are downsampled by a factor of $s$, and superpixel sampling is performed to obtain the corresponding $M$ superpixel features $x_{sp}\in \mathbb{R}^{M\times C}$ and similarity matrix $\textbf{M}_{sim} \in \mathbb{R}^{N\times M}$ (where $N=H\times W$). It is formulated as follows:
\begin{equation}
    x_{in}' = \sigma(\text{Conv}(x_{in}))\downarrow_s, 
\end{equation}
    
\begin{equation}
    x_{sp}, \textbf{M}_{sim} = \text{SPS}(x_{in}'),
\end{equation}
where $\text{SPS}(\cdot)$ denotes the superpixel sampling operation. $\downarrow_s$ represents the downsampling operation by a factor of $s$. The superpixel grouping aggregates spatially proximate features with similar appearances into content-adaptive region representations, while preserving the pixel-to-region correspondence through $\textbf{M}_{sim}$.

State-space modeling is subsequently performed over the compact region sequence $x_{sp}$, rather than over the original $N$ densely represented spatial elements:
\begin{equation}
    x_{sp}' = \text{SSM}(x_{sp}),
\end{equation}
Since the number of region representations $M$ is substantially smaller than the number of dense spatial elements $N$, the principal global interaction is conducted over a much shorter sequence. This reduces redundant sequential computation while allowing long-range dependencies to be established among spatially coherent image regions.

After global region-level modeling, the enhanced region representations need to be propagated back to the dense spatial domain for image reconstruction. To this end, we employ Gumbel-Softmax \cite{gumbel-softmax} to convert the similarity matrix $\textbf{M}_{sim}$ into a differentiable discrete assignment mask $\textbf{M}_{mask}\in \mathbb{R}^{N\times M}$: 
\begin{equation}
    \textbf{M}_{mask} = \text{Gumbel-Softmax}(\textbf{M}_{sim}),
\end{equation}
The enhanced region representations are then projected to their associated spatial locations through the learned pixel-to-region assignments, producing a dense modulation map $\textbf{A} \in \mathbb{R}^{N\times C}$:
\begin{equation}
    \textbf{A} = \text{Sigmoid}(\textbf{M}_{mask}\otimes x_{sp}'),
\end{equation}
where $\otimes$ denotes matrix multiplication. Finally, the projected modulation map is applied to the transformed spatial feature and combined with a residual connection:
\begin{equation}
    x_{out} = (\textbf{A} \odot x_{in}')\uparrow_s+x_{in},
\end{equation}
where $\odot$ denotes element-wise multiplication and $\uparrow_s$ denotes upsampling with scale factor $s$. Consequently, SP-SSM retains dense spatial outputs required for image reconstruction while performing its principal global sequence modeling over compact region-level representations. By explicitly constructing the visual sequence according to spatial coherence and feature similarity, SP-SSM reduces redundant dense sequence modeling and facilitates long-range interactions among structurally related image regions.

% Subsequently, a SSM is employed to perform global information modeling on the superpixel feature $x_{sp}$, yielding an enhanced superpixel feature $x_{sp}'\in \mathbb{R}^{M\times C}$. We use Gumbel-Softmax \cite{gumbel-softmax} to transform the similarity matrix $\textbf{M}_{sim}$ into a differentiable one-hot mask $\textbf{M}_{mask}\in \mathbb{R}^{N\times M}$, while effectively maintaining stable superpixel to pixel propagation during the training process. Then, matrix multiplication followed by a sigmoid function is utilized to derive the global attention feature $\textbf{A} \in \mathbb{R}^{N\times C}$, as follows:
% \begin{align}
%     x_{sp}'&= \text{SSM}(x_{sp}) \\
%     \textbf{M}_{mask} &= \text{Gumbel-Softmax}(\textbf{M}_{sim})\\
%     \textbf{A} &= \text{Sigmoid}(\textbf{M}_{mask}\otimes x_{sp}')
% \end{align}
% where $\text{Sigmoid}(\cdot)$ and $\otimes$ denote sigmoid function and matrix multiplication. The final output of this module is obtained by multiplying the attention feature $\textbf{A}$ with $x_{in}'$, followed by the addition of the original transformed feature $x_{in}$, as follows:
% \begin{equation}
%     x_{out} = (\textbf{A} \odot x_{in}')\uparrow_s+x_{in}
% \end{equation}
% Since superpixels encapsulate comprehensive semantic information, the resulting output features effectively capture correlations among distinct semantics.

\subsubsection{Superpixel Sampling}
\begin{figure}[t]
    \centering
    \includegraphics[width=0.9\columnwidth]{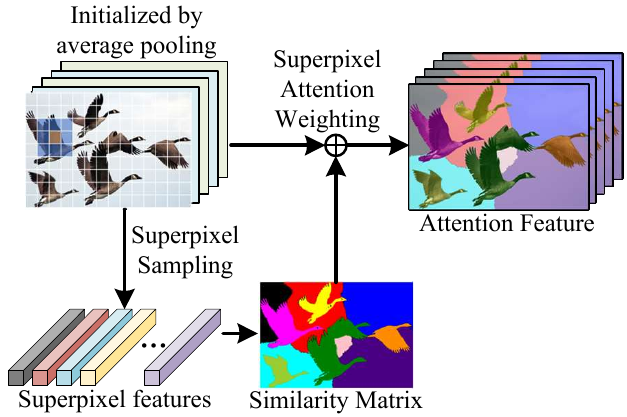}
    % \vspace{-0.2cm}
    \caption{Superpixel sampling of our method, which initializes the superpixel features by average pooling, and then generates the superpixel features and similarity matrix.}
    \vspace{-0.5cm}
    \label{fig:superpixel}
\end{figure}

% We follow the soft $k$-means-based superpixel algorithm in SSN \cite{ssn} to perform superpixel sampling on images.
We adopt the differentiable soft $k$-means-based superpixel formulation of SSN \cite{ssn} to construct content-adaptive region representations. Given an input $\textbf{x}\in \mathbb{R}^{N\times C}$ (where $N=H\times W$), $M$ superpixels $\textbf{s}\in \mathbb{R}^{M\times C}$ and similarity matrix $\textbf{M}_{sim} \in \mathbb{R}^{N \times M}$ are obtained through $T$ iterations, maximizing their association with the corresponding pixels. First, as shown in Fig. \ref{fig:superpixel}, we use average pooling to initialize superpixels $\textbf{s}^{0}$. Then, we conduct iterations using a similarity matrix that calculates the similarity between each pixel and superpixel. It can be formulated as follows:
\begin{equation}
    \textbf{M}_{sim}^t(i,j) = e^{-||\textbf{x}(i)-\textbf{s}^{t-1}(j)||_2^{2}},
\end{equation}
The similarity computation is restricted to neighboring superpixel candidates, preserving local spatial coherence while avoiding global pixel-to-region matching. Subsequently, we can obtain the superpixel $\textbf{s}^t$ by computing a weighted sum of all pixels, defined as:
\begin{equation}
    \textbf{s}^t_j=\frac{1}{\textbf{z}^t(j)}\sum_i \textbf{M}_{sim}^t(i,j)\textbf{x}(i),
\end{equation}
where $\textbf{z}^t(j)=\sum_i \textbf{M}_{sim}^t(i,j)$ denotes the normalization term along the column. After $T$ iterations, the resulting superpixel representations $\textbf{s}^T$ and similarity matrix $\textbf{M}_{sim}^T$ provide a differentiable association between dense spatial features and compact region-level representations. 

Unlike regular grid partitioning, the spatial support and pixel membership of these regions are determined by the input features, allowing the grouping to adapt to local appearance and structural variations. Although the total number of regions $M$ is predefined to maintain a stable computational budget, the resulting spatial partition is content dependent. Therefore, a fixed region budget does not imply a fixed spatial partition. 

The learned pixel-to-region associations are subsequently used by SP-SSM in two complementary directions. First, spatially coherent features are aggregated into compact region representations for global state-space modeling. Second, after the region representations are enhanced by the SSM, the same associations provide a pathway for propagating the region-level information back to the corresponding dense spatial locations. In this way, superpixel sampling serves as the bridge between dense spatial features and the compact region-level sequence employed by SP-SSM.

% Using the similarity matrix, we can assign each pixel to its most similar superpixel, thus generating the corresponding mask, as depicted in Fig. \ref{fig:superpixel}. Therefore, with the superpixels and their respective masks, we can perform superpixel-based attention weighting on pixels across distinct regions. Our proposed SP-SSM utilizes this critical insight by employing a SSM to assign weights to superpixels, enabling weighted processing of semantic information across distinct regions. 

\subsection{Local Spatial Modulating Experts}
MSS-MoE primarily performs long-range interaction over compact region-level representations. While this way improves the efficiency of global modeling, aggregating neighboring spatial features into region representations may attenuate fine spatial variations that are critical for high-quality SR reconstruction. Moreover, region-level sequence modeling emphasizes inter-region interactions, whereas dense local dependencies within and around individual regions are not explicitly modeled by SP-SSM alone. We therefore introduce the Local Spatial Modulating Expert (LSME) as a complementary local refinement branch.

Given the input dense feature representation, LSME employs a Local Mixed Attention (LMA) module that combines shifted-window multi-head self-attention (SwinMHSA) with channel attention, as depicted in Fig. \ref{fig:network}. Channel attention recalibrates informative responses across feature channels, while SwinMHSA captures fine-grained spatial dependencies within local windows. These two operations provide complementary channel-wise and spatial-wise refinement in the dense feature domain. Specifically, channel attention first emphasizes informative feature responses through global channel statistics and adaptive channel reweighting. SwinMHSA subsequently models spatial interactions within shifted local windows, allowing neighboring pixels and fine structures to exchange information without introducing expensive global dense attention. The resulting local representation is further refined by a Gated Feed-Forward Network (GatedFFN).

LSME therefore complements MSS-MoE rather than duplicating its global modeling function. MSS-MoE focuses on efficient long-range interaction over compact region-level sequences, whereas LSME preserves and refines dense local spatial variations that may be weakened during region abstraction. Together, the two branches form a complementary global-region/local-spatial modeling scheme, enabling SP-MoMamba to balance efficient long-range interaction with fine-detail reconstruction.

\begin{table*}[t]
\centering
\caption{Comparison to ultra-lightweight SR models. PSNR (dB $\uparrow$) and SSIM ($\uparrow$) metrics are reported on the Y-channel. The \textbf{best} and \underline{second-best} results are highlighted. GMACs (G) are computed by upscaling to a 1280 × 720 HR image.}
\resizebox{0.9\textwidth}{!}{%
\begin{tabular}{clcccccccccccc}
\hline
                        & \multicolumn{1}{c}{}                                 &                              &                            & \multicolumn{2}{c}{Set5}                                                                                                       & \multicolumn{2}{c}{Set14}                                                                                                      & \multicolumn{2}{c}{BSD100}                                                                                                     & \multicolumn{2}{c}{Urban100}                                                                                                   & \multicolumn{2}{c}{Manga109}                                                                                                   \\
\multirow{-2}{*}{Scale} & \multicolumn{1}{c}{\multirow{-2}{*}{Model}}          & \multirow{-2}{*}{Params $\downarrow$} & \multirow{-2}{*}{GMACs$\downarrow$}    & PSNR$\uparrow$                                                          & SSIM$\uparrow$                                                           & PSNR$\uparrow$                                                          & SSIM$\uparrow$                                                           & PSNR$\uparrow$                                                          & SSIM$\uparrow$                                                           & PSNR$\uparrow$                                                          & SSIM$\uparrow$                                                           & PSNR$\uparrow$                                                          & SSIM$\uparrow$                                                           \\
\hline
\end{tabular}%
 }

\label{tab:sotacp1}
% \vspace{-0.2cm}
\end{table*}

\section{Experiments}\label{exp}

\subsection{Experimental Settings}
\label{sec:exp_set}
\subsubsection{Datasets and Evaluation}
Following standard SR settings \cite{swinir, seemore}, we train our models on DIV2K \cite{DIV2K} and Flickr2K \cite{EDSR} and evaluate them on five widely used benchmark datasets: Set5 \cite{Set5}, Set14 \cite{Set14}, BSD100 \cite{B100}, Urban100 \cite{Urban100}, and Manga109 \cite{manga109}. Reconstruction quality is evaluated using PSNR and SSIM on the Y channel of the YCbCr color space.

% Following the previous SR methods \cite{swinir, seemore}, we utilize two widely-used datasets, DIV2K \cite{DIV2K} and Flickr2K \cite{EDSR} for training. We assess our method performance on five classical benchmark datasets for SR,  Set5 \cite{Set5}, Set14 \cite{Set14}, BSD100 \cite{B100}, Urban100 \cite{Urban100}, and Manga109 \cite{manga109}. We also quantify the effectiveness of our method using the PSNR and SSIM metrics on the Y-channel from the YCbCr color space.

\subsubsection{Implementation Details}
During training, LR images are randomly cropped into $64\times 64$ patches and augmented by random rotations and horizontal/vertical flips. Following \cite{shufflemixer}, we optimize the network using Adam \cite{adam} with an $L_1$ reconstruction objective in both the spatial and frequency domains for 500K iterations. The batch size is set to 32, and the initial learning rate is $1\times 10^{-3}$, which is halved at 250K, 400K, 450K, and 475K iterations. All models are implemented in PyTorch and trained on a single NVIDIA RTX 4090 GPU. We construct two model variants to evaluate SP-MoMamba under different efficiency budgets. SP-MoMamba-T contains 3 LoEs with a channel dimension of 36, whereas SP-MoMamba-B contains 4 LoEs with a channel dimension of 48. Each MSS-MoE contains 3 scale-specific SP-SSM experts with scale factors s=[1,2,4], and top-$1$ routing is adopted during inference. We additionally report SP-MoMamba-B+, which applies the self-ensemble strategy \cite{EDSR} to SP-MoMamba-B. Further details are provided in Supplementary Material.

% To thoroughly train the proposed model, we augment the training data by randomly cropping it into $64\times 64$ patches and further augment it through random rotations, horizontal and vertical flips. Consistent with \cite{shufflemixer}, we use the Adam \cite{adam} optimizer to minimize the $L_1$ norm between the SR output and the HR ground truth in both pixel and frequency domains across 500K iterations. The batch size is set to 32, with an initial learning rate of $1\times 10^{-3}$ which is halved at iterations [250K, 400K, 450K, 475K]. All experiments are implemented using the PyTorch framework and conducted on a single RTX 4090 GPU. We have designed two variants of the SP MoMamba model with different parameter configurations, including the our small model (SP-MoMamba-T) with 3 LoEs and 2 SGMEs and LSMEs. Additionally, our base model (SP-MoMamba-B) with 4 LoEs and 2 SGMEs and LSMEs. The channel dimension of SP-MoMamba-B is 48. For all MSS-MoE modules, we configure three experts with downsampling factors of [1, 2, 4]. To further improve the performance of SP-MoMamba-B, we adopt the self-ensemble strategy \cite{EDSR} to SP-MoMamba-B and denote the self-ensemble version as SP-MoMamba-B+. Further details are provided in Supplementary Material.

\begin{figure*}[t]
    \centering
    \includegraphics[width=\textwidth]{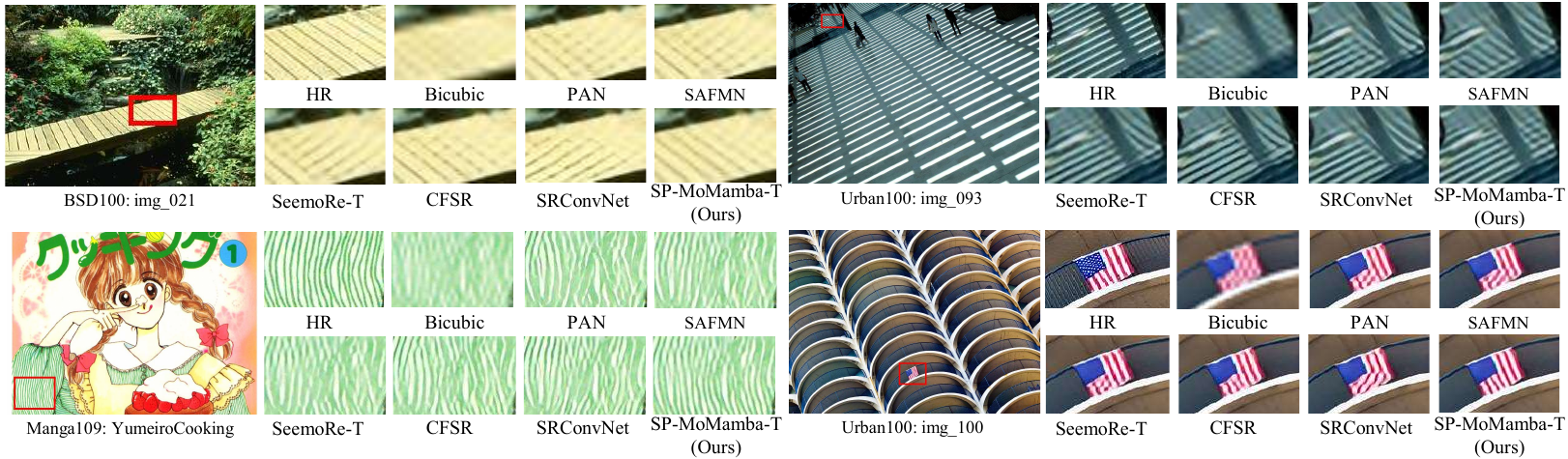}
    % \vspace{-0.5cm}
    \caption{Qualitative comparison of our SP-MoMamba-T with state-of-the-art ultra-lightweight methods on $4\times$ SR.}
    \label{fig:cpsota_ultra}
    \vspace{-0.5cm}
\end{figure*}

\subsection{Comparisons with Ultra-lightweight SR Methods}
\subsubsection{Quantitative comparison}
Following \cite{seemore}, we consider models with fewer than 500K parameters as ultra-lightweight SR models. We compare SP-MoMamba-T with representative efficient approaches, including PAN \cite{pan}, DRSAN \cite{drsan}, SAFMN \cite{SAFMN}, FECAN-tiny \cite{fecan}, SeemoRe-T \cite{seemore}, SRConvNet \cite{srconvnet}, and CFSR \cite{cfsr}.

As shown in Table \ref{tab:sotacp1}, SP-MoMamba-T achieves the best PSNR/SSIM results among the compared methods across the evaluated scaling factors and datasets. The improvement is particularly evident on Urban100 and Manga109, which contain abundant repetitive structures and high-frequency patterns. For example, for $\times 4$ SR on Urban100, SP-MoMamba-T achieves 26.40 dB, compared with 26.23 dB for SeemoRe-T and 26.21 dB for CFSR. On Manga109, it reaches 31.01 dB under the same setting. Importantly, these improvements are obtained with only 271K parameters and 22G GMACs for $\times 4$ SR. The results indicate that performing global interaction over compact region-level representations can provide strong reconstruction capability even under a highly constrained parameter budget. This observation is consistent with the design objective of SP-SSM: reducing redundant dense sequence modeling while retaining effective long-range structural interaction.

\subsubsection{Qualitative comparison}
Fig. \ref{fig:cpsota_ultra} presents visual comparisons for ×4 SR under the ultra-lightweight setting. SP-MoMamba-T shows a consistent advantage in reconstructing structured high-frequency patterns. For example, in the striped regions of img\_044 and the fine flag structures in MomoyamaHaikagura, several competing methods produce blurred or distorted line patterns, whereas SP-MoMamba-T recovers more continuous structures with orientations closer to the HR reference. Similar behavior can be observed in the geometric patterns of img\_076, where our method better preserves line continuity and edge sharpness. These examples suggest that the proposed region-level global modeling remains effective for complex structured content despite the compact model size, while the complementary local refinement branch further contributes to recovering fine spatial details.

\begin{table*}[]
\centering
\caption{Comparison to standard lightweight SR models. PSNR (dB $\uparrow$) and SSIM ($\uparrow$) metrics are reported on the Y-channel. The \textbf{best} and \underline{second-best} results are highlighted. GMACs (G) are computed by upscaling to a 1280 × 720 HR image. * denotes results obtained by retraining under an equivalent experimental configuration.}
\resizebox{0.9\textwidth}{!}{%
\begin{tabular}{clcccccccccccc}
\hline
                        & \multicolumn{1}{c}{}                                 &                              &                             & \multicolumn{2}{c}{Set5}                                                                                     & \multicolumn{2}{c}{Set14}                                                                                    & \multicolumn{2}{c}{BSD100}                                                                                   & \multicolumn{2}{c}{Urban100}                                                                                 & \multicolumn{2}{c}{Manga109}                \\                                                      
\multirow{-2}{*}{Scale} & \multicolumn{1}{c}{\multirow{-2}{*}{Model}}          & \multirow{-2}{*}{Params $\downarrow$} & \multirow{-2}{*}{GMACs$\downarrow$}     & PSNR$\uparrow$                                                 & SSIM$\uparrow$                                                  & PSNR$\uparrow$                                                 & SSIM$\uparrow$                                                  & PSNR$\uparrow$                                                 & SSIM$\uparrow$                                                  & PSNR$\uparrow$                                                 & SSIM$\uparrow$                                                  & PSNR$\uparrow$                                                 & SSIM$\uparrow$    \\
\hline
\end{tabular}%
}
\label{tab:sotacp2}
\end{table*}

\begin{figure*}[ht]
    \centering
    \includegraphics[width=\textwidth]{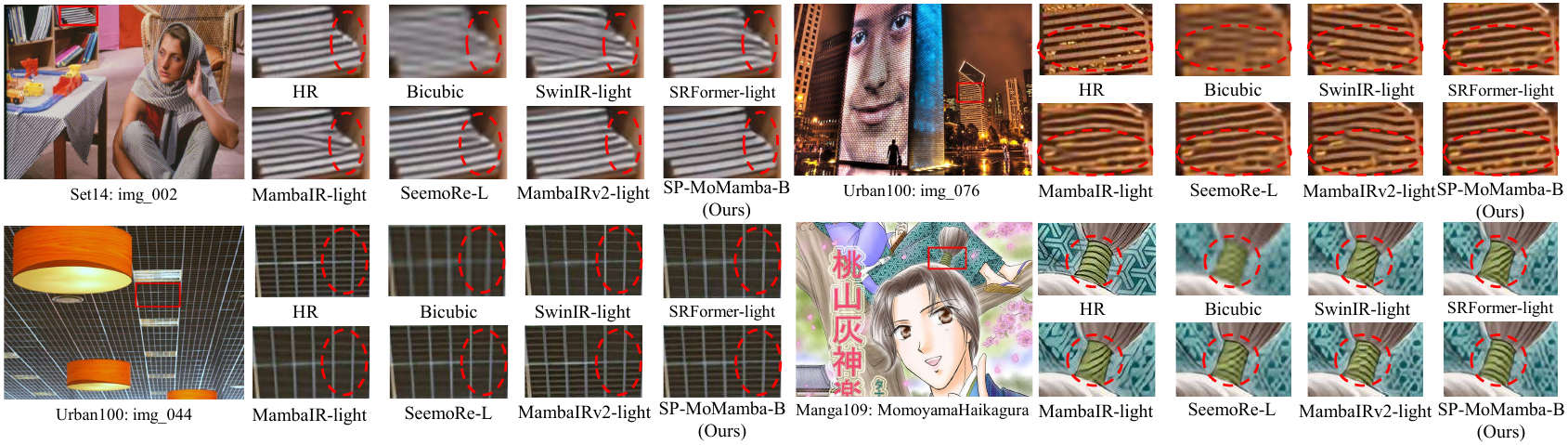}
    \vspace{-0.3cm}
    \caption{Qualitative comparison of our SP-MoMamba-B with state-of-the-art methods on $4\times$ SR.}
    \label{fig:cpsota}
\end{figure*}

\begin{figure*}[ht]
    \centering
    \includegraphics[width=0.8\textwidth]{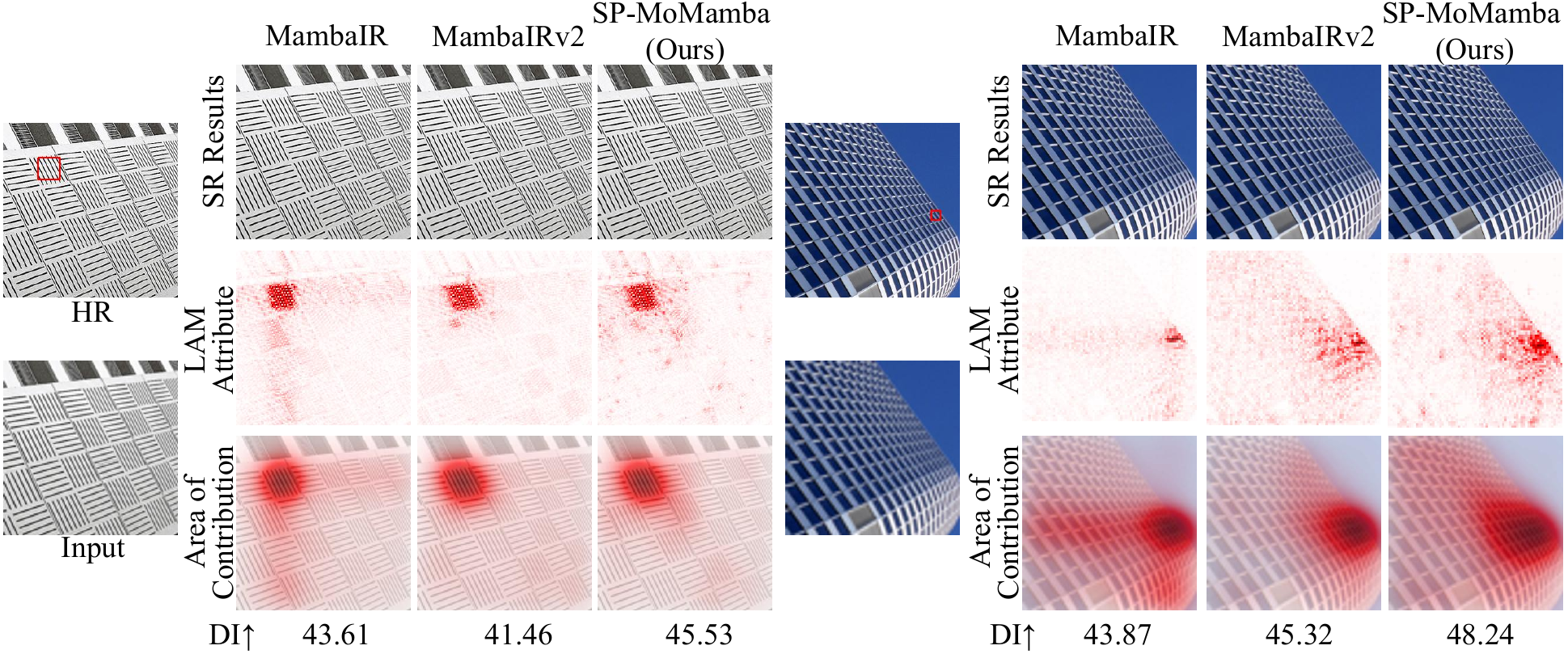}
    % \vspace{-0.5cm}
    \caption{Comparison of Local Attribution Maps (LAM) \cite{LAM} and Diffusion Indices (DI). The red dot denotes the target pixel used for attribution analysis. SP-MoMamba exhibits a broader spatial contribution range and higher DI scores in the illustrated examples, indicating increased utilization of spatially distant information during reconstruction.}
    \label{lamvisual}

\end{figure*}

\begin{figure*}[t]
    \centering
    \includegraphics[width=\textwidth]{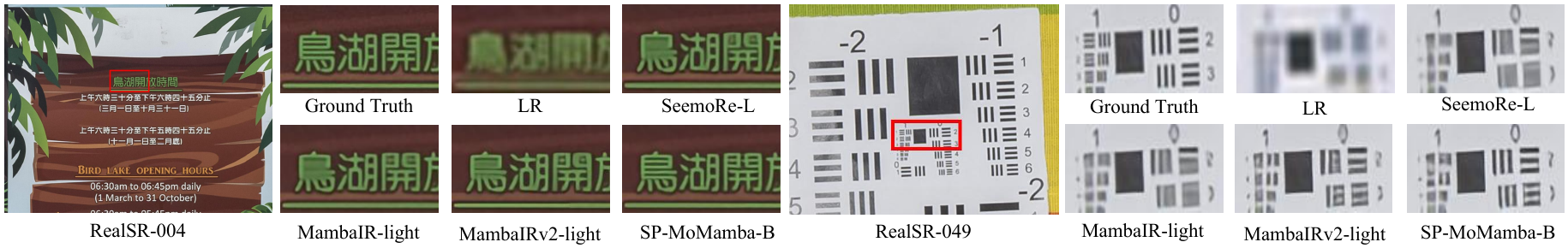}
    % \vspace{-0.2cm}
    \caption{Qualitative comparisons with existing efficient SR methods on the real-world test set RealSRv3 \cite{realsr}.}
    \label{fig:realsr}
\end{figure*}

\subsection{Comparisons with lightweight SR Methods}
\subsubsection{Quantitative comparison}
We further evaluate SP-MoMamba-B and SP-MoMamba-B+ under the standard lightweight SR setting against representative CNN-based models (SPIN \cite{SPIN}, CAMixerSR \cite{camixer}, FECAN-light \cite{fecan}, SeemoRe-L \cite{seemore}, CRAFT \cite{CRAFT}, and DMNet \cite{DMNet}), Transformer-based architectures (e.g., SwinIR-Light \cite{swinir}, SRFormer-Light \cite{srformer}), and recent advanced Mamba-based methods (e.g., MambaIR-light \cite{mambair}, MambaIRv2-light \cite{mambairv2}), as summarized in Table \ref{tab:sotacp2}. 

SP-MoMamba-B provides a favorable balance between model size and reconstruction accuracy. For ×4 SR, the model contains 559K parameters and requires 46G GMACs, while achieving 32.56 dB, 28.93 dB, 27.78 dB, 26.76 dB, and 31.51 dB on Set5, Set14, BSD100, Urban100, and Manga109, respectively. The self-ensemble variant SP-MoMamba-B+ further improves reconstruction accuracy, reaching 31.75 dB on Manga109. The advantage is particularly evident on datasets dominated by structured and repetitive textures. For example, SP-MoMamba-B achieves 31.51 dB on Manga109, compared with 31.48 dB for SeemoRe-L and 31.24 dB for MambaIRv2-light. These results indicate that the proposed region-level sequence modeling provides competitive reconstruction accuracy with a relatively compact model and moderate computational cost. Overall, rather than consistently optimizing a single metric, SP-MoMamba is designed to achieve a favorable performance–complexity trade-off, which is further examined through runtime and memory measurements in Sec. \ref{ve}.

\subsubsection{Qualitative comparison}
Fig. \ref{fig:cpsota} compares SP-MoMamba-B with representative lightweight methods on challenging structured regions. On img\_044 and img\_076 from Urban100, our method reconstructs more continuous grid and line structures, while several competing approaches exhibit blurred intersections or distorted orientations. A similar advantage can be observed on the dense repetitive patterns of MomoyamaHaikagura from Manga109, where SP-MoMamba-B preserves finer structural regularity. 

To further examine how different models utilize spatial information during reconstruction, we analyze Local Attribution Maps (LAM) \cite{LAM} and the Diffusion Index (DI) in Fig. \ref{lamvisual}. MambaIR and MambaIRv2 exhibit relatively localized attribution patterns, whereas SP-MoMamba activates a broader set of spatially separated regions. For the two examples, SP-MoMamba obtains DI scores of 45.53 and 48.24, higher than the corresponding scores of the compared Mamba-based models. The broader attribution range indicates that region-level sequence modeling facilitates the utilization of spatially distant structural information during reconstruction. Together with the visual comparisons, these observations provide empirical support for the intended long-range interaction capability of SP-SSM, without requiring dense global sequence modeling.

\subsection{Evaluation on Real-World Super-Resolution}
To evaluate the applicability of SP-MoMamba beyond synthetic bicubic degradation, we conduct ×4 SR experiments on the RealSRv3 dataset \cite{realsr}. As reported in Table \ref{tab:realsr}, SP-MoMamba-B achieves 29.31 dB PSNR, 0.8312 SSIM, and 0.2711 LPIPS, outperforming the compared lightweight models under all three reconstruction metrics. Meanwhile, it uses 559K parameters and 46G GMACs, which are lower than those of the compared MambaIR and MambaIRv2 models.

Fig. \ref{fig:realsr} further provides qualitative comparisons on real-world images. In RealSR-004, SP-MoMamba-B recovers clearer character strokes, while in RealSR-049 it reconstructs more distinguishable parallel line patterns with reduced aliasing. These results suggest that the proposed region-level modeling framework generalizes beyond synthetic SR and remains effective for structured content under more complex real-world degradations.

Overall, the RealSRv3 results provide additional evidence that SP-MoMamba maintains a favorable balance between reconstruction quality and computational cost under practical degradation conditions.

\begin{table*}[t]
\centering
\caption{Quantitative evaluation of our method and existing lightweight SR methods on a real-world test set.}
\resizebox{0.7\textwidth}{!}{%
\begin{tabular}{cl|cc|c}
\hline
\multicolumn{1}{l}{{\color[HTML]{1F1F1F} }}                                 & {\color[HTML]{1F1F1F} }                                                  & {\color[HTML]{1F1F1F} }                                  & {\color[HTML]{1F1F1F} }                                 & {\color[HTML]{1F1F1F} \textbf{RealSRv3 {\cite{realsr}}}}   \\
\multicolumn{1}{l}{\multirow{-2}{*}{{\color[HTML]{1F1F1F} \textbf{Scale}}}} & \multirow{-2}{*}{{\color[HTML]{1F1F1F} \textbf{Method}}}                & \multirow{-2}{*}{{\color[HTML]{1F1F1F} \textbf{Params}}} & \multirow{-2}{*}{{\color[HTML]{1F1F1F} \textbf{GMACs}}} & {\color[HTML]{1F1F1F} \textbf{PSNR$\uparrow$/SSIM$\uparrow$/LPIPS$\downarrow$}}  \\ \hline
\multicolumn{1}{c|}{{\color[HTML]{1F1F1F} }}                                & {\color[HTML]{1F1F1F} SwinIR-Light \cite{swinir}}                              & {\color[HTML]{1F1F1F} 930K}                              & {\color[HTML]{1F1F1F} 64G}                              & {\color[HTML]{1F1F1F} 29.18/0.8278/0.2756}          \\
\multicolumn{1}{c|}{{\color[HTML]{1F1F1F} }}                                & {\color[HTML]{1F1F1F} MambaIR-light \cite{mambair}}                                   & {\color[HTML]{1F1F1F} 924K}                              & {\color[HTML]{1F1F1F} 85G}                              & {\color[HTML]{1F1F1F} 29.22/0.8286/0.2748}          \\
\multicolumn{1}{c|}{{\color[HTML]{1F1F1F} }}                                & {\color[HTML]{1F1F1F} SeemoRe-L \cite{seemore}}                                         & {\color[HTML]{1F1F1F} 715K}                              & {\color[HTML]{1F1F1F} 50G}                              & {\color[HTML]{1F1F1F} 29.24/0.8293/0.2764}          \\
\multicolumn{1}{c|}{{\color[HTML]{1F1F1F} }}                                & {\color[HTML]{1F1F1F} MambaIRv2-light \cite{mambairv2}}                                 & 790K                                                     & 76G                                                     & 29.27/0.8295/0.2738                                 \\
\multicolumn{1}{c|}{\multirow{-5}{*}{{\color[HTML]{1F1F1F} $\times$4}}}           & \multicolumn{1}{c|}{{\color[HTML]{1F1F1F} \textbf{SP-MoMamba-B (Ours)}}} & {\color[HTML]{1F1F1F} \textbf{559K}}                     & {\color[HTML]{1F1F1F} \textbf{46G}}                     & {\color[HTML]{1F1F1F} \textbf{29.31/0.8312/0.2711}} \\ \hline
\end{tabular}%
}
\label{tab:realsr}
\end{table*}

\subsection{Comparison of Model Efficiency} 
\label{ve}
\begin{figure}[t]
    \centering
    \includegraphics[width=0.9\columnwidth]{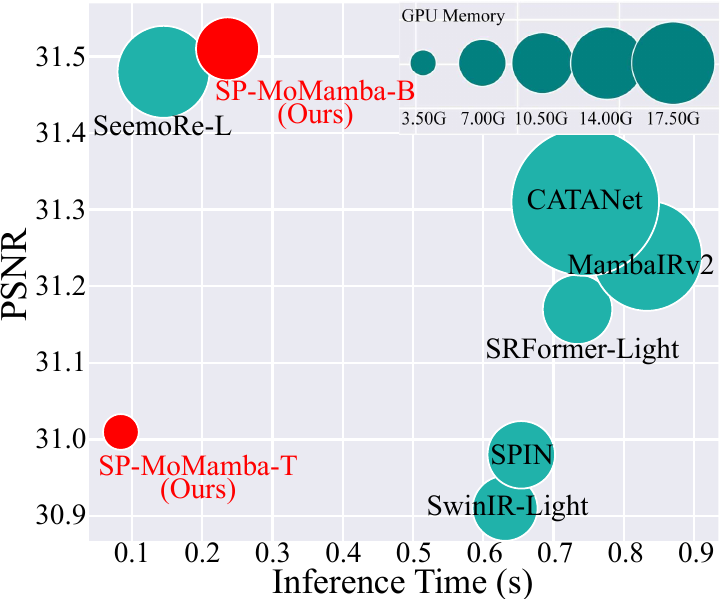}
    \caption{Comparison between performance vs Inference times and GPU Memory on Manga109 ×4 dataset. Inference times and GPU Memory are calculated on 720p HR image.}
    \label{fig:efficient}
\end{figure}

A central motivation of SP-SSM is to avoid performing the principal global interaction over unnecessarily long dense visual sequences. We therefore evaluate not only theoretical complexity but also practical inference latency and GPU memory consumption. All measurements in Fig. \ref{fig:efficient} and Table \ref{tab:complexity} are obtained for $\times4$ SR with a 720p HR output on Manga109 under the same evaluation environment.

SP-MoMamba-T achieves the lowest inference latency and GPU memory consumption among the compared methods, requiring only 0.084 s and 4258.1 MB while obtaining a PSNR of 31.01 dB. SP-MoMamba-B improves reconstruction quality to 31.51 dB while maintaining a latency of 0.236 s and memory consumption of 6572.9 MB. In comparison, MambaIRv2-light requires 0.833 s and 13652.3 MB, whereas CATANet consumes 21962 MB of GPU memory. Compared with the efficient SeemoRe-L baseline, SP-MoMamba-B improves PSNR from 31.48 dB to 31.51 dB while reducing GPU memory from 10464.7 MB to 6572.9 MB. These practical measurements support the computational motivation of the proposed region-level formulation: conducting the principal global interaction over compact region representations can reduce runtime and memory overhead while maintaining competitive reconstruction accuracy. As visualized in Fig. 10, the two SP-MoMamba variants occupy different operating points on the performance-efficiency spectrum: SP-MoMamba-T prioritizes aggressive efficiency, whereas SP-MoMamba-B provides higher reconstruction accuracy under a moderate computational budget.

\begin{table}[t]
\centering
\caption{Comparison between performance vs Inference times and GPU Memory on Manga109 ×4 dataset.}
% \resizebox{0.9\columnwidth}{!}{%
\begin{tabular}{lccc}
\hline
Method          & Time (s)       & GPU Memory      & PSNR           \\ \hline
SwinIR-Light    & 0.631          & 6802.7          & 30.91          \\
SRFormer-Light  & 0.734          & 7319.4          & 31.17          \\
SPIN            & 0.654          & 7083            & 30.98          \\
Mambairv2-Light & 0.833          & 13652.3         & 31.24          \\
CATANet         & 0.745          & 21962           & 31.31          \\
SeemoRe-L       & {0.145} & 10464.7         & 31.48          \\ \hline
SP-MoMamba-T    & \textbf{0.084}          & \textbf{4258.1} & 31.01          \\
SP-MoMamba-B    & 0.236          & 6572.9          & \textbf{31.51} \\ \hline
\end{tabular}%
% }
\label{tab:complexity}
\vspace{-0.2cm}

\end{table}

\subsection{Ablation Study}
We conduct ablation studies on SP-MoMamba-T under the $\times4$ setting to examine three aspects of the proposed framework: (a) whether region-level sequence modeling provides advantages over dense SSM modeling, (b) whether adaptive representation granularity is beneficial for MSS-MoE, and (c) whether local spatial refinement complements region-level global interaction. 
% More ablation studies are in supplementary materials.

\begin{table*}[ht]
\centering
\caption{Ablation on key components of SP-MoMamba. We show PSNR results for $\times$4 upscaling.}
\resizebox{\textwidth}{!}{%
\begin{tabular}{c|ccc|c|c|ccrr}
\hline
  & \multicolumn{3}{c|}{SGME} & \multicolumn{1}{c|}{LSME} & FFN & \multicolumn{4}{c}{Performance}  \\
\multicolumn{1}{l|}{{\color[HTML]{1F1F1F} Method}}     & {SSM} & SPS                      & MoE                      & LMA                      & {\color[HTML]{1F1F1F} GatedFFN} & {\color[HTML]{1F1F1F} Params} & {\color[HTML]{1F1F1F} GMACs} & \multicolumn{1}{l}{{\color[HTML]{1F1F1F} BSD100}} & \multicolumn{1}{l}{{\color[HTML]{1F1F1F} Urban100}} \\ \hline
\multicolumn{1}{l|}{{\color[HTML]{1F1F1F} Baseline}}   & {\color[HTML]{1F1F1F} -}   & {\color[HTML]{1F1F1F} -} & {\color[HTML]{1F1F1F} -} & {\color[HTML]{1F1F1F} -} & {\color[HTML]{1F1F1F} -}        & {\color[HTML]{1F1F1F} 189K}   & {\color[HTML]{1F1F1F} 8.4}   & {\color[HTML]{1F1F1F} 27.42}                      & {\color[HTML]{1F1F1F} 25.55}                        \\
{\color[HTML]{1F1F1F} }                               & {\color[HTML]{1F1F1F} \ding{52}}   & {\color[HTML]{1F1F1F} -} & {\color[HTML]{1F1F1F} -} & {\color[HTML]{1F1F1F} -} & {\color[HTML]{1F1F1F} -}        & 230K                          & 18.7                         & {\color[HTML]{1F1F1F} 27.48}                      & {\color[HTML]{1F1F1F} 25.64}                        \\
{\color[HTML]{1F1F1F} }                               & {\color[HTML]{1F1F1F} \ding{52}}   & {\color[HTML]{1F1F1F} \ding{52}} & -                        & -                        & {\color[HTML]{1F1F1F} -}        & 197K                          & 16.7                         & {\color[HTML]{1F1F1F} 27.54}                      & {\color[HTML]{1F1F1F} 25.86}                        \\
{\color[HTML]{1F1F1F} }                               & {\color[HTML]{1F1F1F} \ding{52}}   & {\color[HTML]{1F1F1F} \ding{52}} & {\color[HTML]{1F1F1F} \ding{52}} & -                        & {\color[HTML]{1F1F1F} -}        & {\color[HTML]{1F1F1F} 242K}   & {\color[HTML]{1F1F1F} 16.9}  & {\color[HTML]{1F1F1F} 27.59}                      & {\color[HTML]{1F1F1F} 26.07}                        \\
{\color[HTML]{1F1F1F} }                               & -                          & -                        & -                        & {\color[HTML]{1F1F1F} \ding{52}} & {\color[HTML]{1F1F1F} \ding{52}}        & 211K                          & 16.6                         & {\color[HTML]{1F1F1F} 27.61}                      & {\color[HTML]{1F1F1F} 26.14}                        \\
{\color[HTML]{1F1F1F} }                               & {\color[HTML]{1F1F1F} \ding{52}}   & {\color[HTML]{1F1F1F} \ding{52}} & {\color[HTML]{1F1F1F} \ding{52}} & -                        & {\color[HTML]{1F1F1F} \ding{52}}        & {\color[HTML]{1F1F1F} 261K}   & {\color[HTML]{1F1F1F} 19.8}  & {\color[HTML]{1F1F1F} 27.63}                      & {\color[HTML]{1F1F1F} 26.14}                        \\
\multirow{-6}{*}{{\color[HTML]{1F1F1F} SP-MoMamba-T}} & {\color[HTML]{1F1F1F} \ding{52}}   & {\color[HTML]{1F1F1F} \ding{52}} & {\color[HTML]{1F1F1F} \ding{52}} & {\color[HTML]{1F1F1F} \ding{52}} & {\color[HTML]{1F1F1F} \ding{52}}        & {\color[HTML]{1F1F1F} 271K}   & {\color[HTML]{1F1F1F} 22}    & {\color[HTML]{1F1F1F} \textbf{27.69}}                      & {\color[HTML]{1F1F1F} \textbf{26.40}}                        \\ \hline
\end{tabular}%
}
\label{tab:ab1}
\end{table*}

\begin{figure*}[ht]
    \centering
    \includegraphics[width=\textwidth]{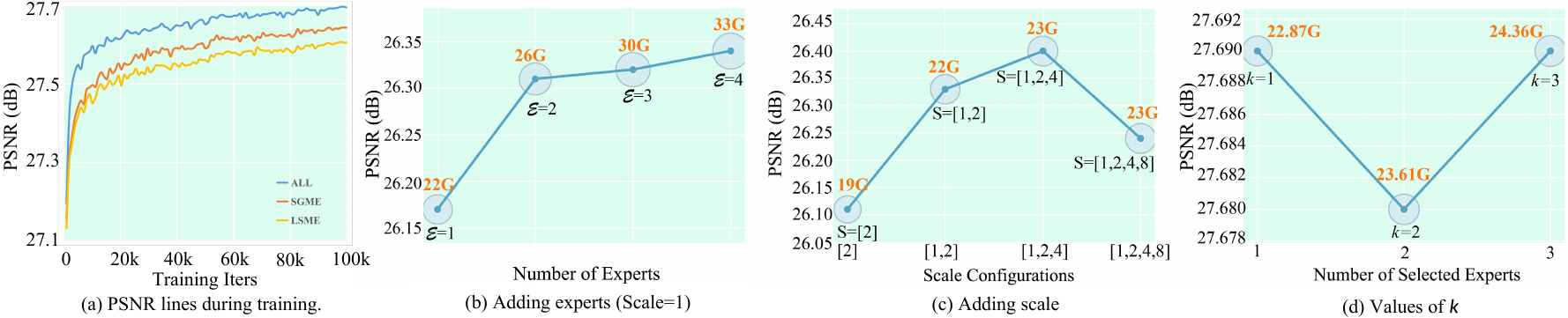}
    \caption{Ablation studies of SP-MoMamba-T for $\times4$ SR. (a) Training PSNR curves of different macro-architectural configurations. (b) Effect of the number of same-scale experts. (c) Effect of different region-level modeling granularities. (d) Effect of the number $k$ of activated experts in MSS-MoE.}
    \label{fig:abstudy}
\end{figure*}

\begin{table}[t]
\centering
\caption{Superpixel sampling method. We compared the performance impact of different superpixel sampling method separately.}
\begin{tabular}{lcccc}
\hline
Method       & Params & GMACs & BSD100 & Urban100 \\ \hline
w SLIC \cite{slic}         & 271K   & 26    & 27.51  & 26.03    \\
w STA \cite{sta}          & 460K   & 38    & 27.72  & 26.49    \\
w SSN (Ours)   & 271K   & 22    & 27.69  & 26.30     \\ \hline
\end{tabular}
\label{spab}
\vspace{-0.4cm}
\end{table}

\begin{table}[t]
\centering

\footnotesize
\setlength{\tabcolsep}{2pt}
\renewcommand{\arraystretch}{1.08}

\caption{Analysis of the impact of the number of superpixels on the performance of MSS-MoE.}
% \resizebox{\columnwidth}{!}{%
\begin{tabular}{@{}cccccc@{}}
\toprule
\multicolumn{1}{l}{Method}  &Experts    & Superpixels & GMACs & BSD100 & Urban100 \\ \midrule
\multirow{3}{*}{SP-MoMmaba-T}  & 3 & \textbf{{[}64,64,64{]}}  & {22}    & \textbf{27.69}  & \textbf{26.40} \\
                               & 3  & {{[}16,32,64{]}}  & \textbf{21}   & {27.65}   & 26.28\\
                               & 3  & {[}32,64,128{]} & 39    & 27.66   & 26.30 \\ \bottomrule
\end{tabular}%
% }
\vspace{-0.4cm}
\label{tab:numsp}
\end{table}

\begin{figure*}[t]
    \centering
    \includegraphics[width=1\textwidth]{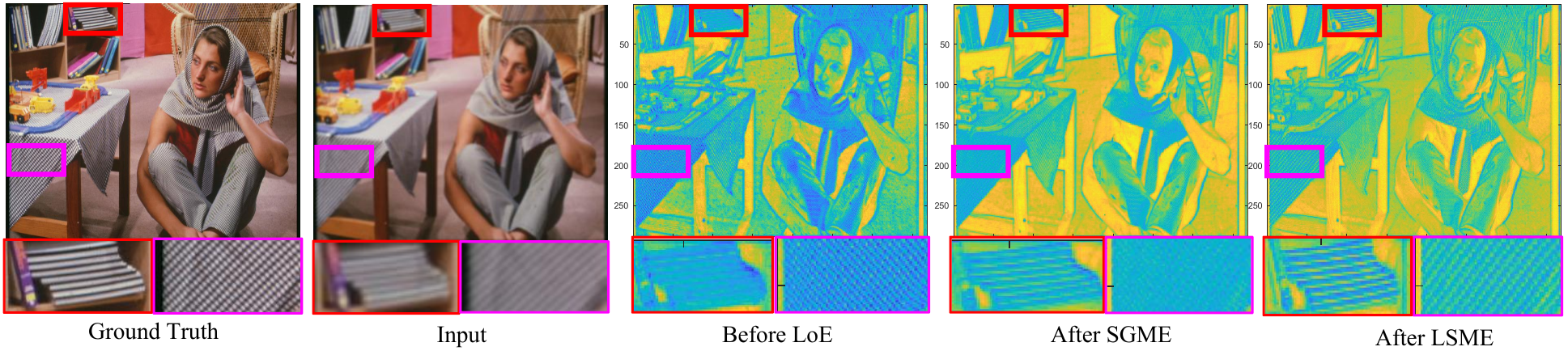}
    % \vspace{-0.2cm}
    \caption{Visualization of intermediate features before and after SGME and LSME. SGME enhances large-scale structural responses, while LSME further refines localized edge and texture representations.}
    \label{fig:visualization}
    % \vspace{-0.5cm}
\end{figure*}

\subsubsection{Macro Architecture Components}
Table \ref{tab:ab1} evaluates the major components of SP-MoMamba by progressively introducing them into a residual-block baseline. Adding a conventional SSM improves PSNR from 27.42/25.55 dB to 27.48/25.64 dB on BSD100/Urban100, while increasing computation from 8.4G to 18.7G GMACs. Introducing Superpixel Sampling (SPS) further improves the results to 27.54/25.86 dB while reducing the cost to 16.7G GMACs, indicating that compact region-level sequence modeling improves both reconstruction accuracy and computational efficiency over dense SSM modeling. Building on SP-SSM, the MoE design further raises Urban100 PSNR to 26.07 dB, supporting the benefit of multiple region-level modeling granularities. The complete SP-MoMamba-T achieves 27.69 dB on BSD100 and 26.40 dB on Urban100, corresponding to gains of 0.27 dB and 0.85 dB over the baseline, respectively.

Fig. \ref{fig:abstudy}(a) further shows that the complete architecture maintains a higher PSNR trajectory during training and converges to better final performance than the partial configurations. Fig. \ref{fig:visualization} provides complementary feature-level evidence: SGME strengthens responses around large-scale structural patterns, while the subsequent LSME enhances localized edge and texture representations. Together, these results verify the complementary roles of region-level global modeling, adaptive granularity, and dense local refinement.

\subsubsection{Representation Granularity and Routing in MSS-MoE}
We next investigate whether the improvement of MSS-MoE originates merely from increasing the number of experts or from modeling multiple region granularities. Fig. \ref{fig:abstudy}(b) first increases the number of experts while fixing the SP-SSM scale to $s=1$. Although additional same-scale experts provide moderate performance improvements, the computational cost increases rapidly, reaching approximately 33G GMACs with four experts. In contrast, Fig. \ref{fig:abstudy}(c) varies the scale composition of the experts. The multi-scale configuration $s=[1,2,4]$ achieves the best performance-efficiency balance, reaching 26.40 dB with substantially lower computation than the four-expert single-scale configuration. This comparison suggests that the benefit of MSS-MoE does not arise solely from increased expert capacity; diversifying the region-level representation granularity is also important for handling heterogeneous image structures.

Expanding the scale set to $s=[1,2,4,8]$ reduces PSNR to 26.24 dB. Rather than attributing this degradation to “semantic information loss,” we observe that excessive downsampling produces overly coarse region representations, which can suppress spatial details required for SR reconstruction. We therefore adopt $s=[1,2,4]$ as the default expert configuration. We further evaluate the number of activated experts $k$ in Fig. 11(d). Increasing $k$ from 1 to 3 raises the computation from 22.87G to 24.36G GMACs without improving PSNR, while $k=2$ even produces a slight degradation. Therefore, top-$1$ routing is adopted in the final model. These results indicate that activating multiple granularities simultaneously is unnecessary and selecting a single input-dependent granularity provides a better performance-efficiency trade-off under the evaluated setting.

\begin{figure}[t]
\centering
\resizebox{0.75\columnwidth}{!}{%
\begin{tabular}{ccc}
\hline
{\color[HTML]{1F1F1F} Aggregation method} & {\color[HTML]{1F1F1F} Urban100} & {\color[HTML]{1F1F1F} Manga109} \\ \hline
 {\color[HTML]{1F1F1F} (a)} & {\color[HTML]{1F1F1F} 32.11}    & {\color[HTML]{1F1F1F} 38.96}    \\
 {\color[HTML]{1F1F1F} (b)}              & {\color[HTML]{1F1F1F} 32.15}    & {\color[HTML]{1F1F1F} 38.99}    \\
 {\color[HTML]{1F1F1F} (c)}              & {\color[HTML]{1F1F1F} 32.17}    & {\color[HTML]{1F1F1F} 38.97}    \\
 {\color[HTML]{1F1F1F} (d)}   & {\color[HTML]{1F1F1F} 32.22}    & {\color[HTML]{1F1F1F} 39.01}    \\ \hline
\end{tabular}
}

\includegraphics[width=0.85\columnwidth]{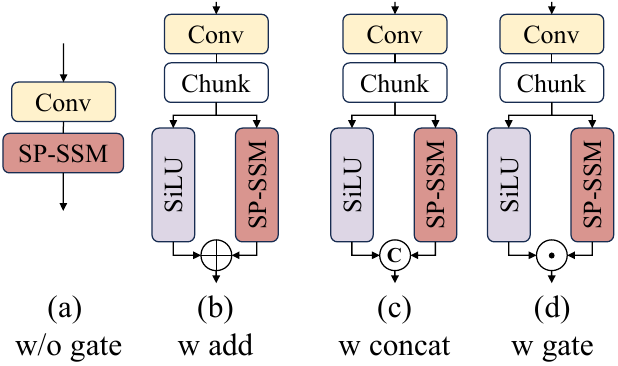}

\caption{Analysis of gating mechanism on SP-SSM. We provide further insights in the design decisions of our SP-MoMamba-T framework for ×2 upscaling.}
\vspace{-0.4cm}
\label{fig:spssm}
\end{figure}

\begin{table}[t]
    \centering
    \caption{Ablation on the Local Mixed Attention (LMA) mechanism. We show results for $\times4$ upscaling.}
    \begin{tabular}{@{}l|cccc@{}}
    \toprule
    Method    & Params & \multicolumn{1}{c}{GMACs} & BSD100         & Urban100      \\ \midrule
    SwinMHSA & 224K   & 20                        & 27.67          & 26.28         \\
    CA        & 240K   & 13                        & 27.65         & 26.22         \\
    LMA       & 271K   & 22                        & \textbf{27.69} & \textbf{26.40} \\ \bottomrule
    \end{tabular}%
    \label{tab:ab3}
    \vspace{-0.4cm}
\end{table}

\subsubsection{Micro-Architectural Design Choices}
We further investigate several internal design choices of SP-MoMamba, including the superpixel construction strategy, region budget, gating mechanism, and local refinement design, as detailed in Tables \ref{spab}, \ref{tab:numsp}, \ref{tab:ab3} and Fig. \ref{fig:spssm}.

\begin{figure*}[ht]
  \centering
  \includegraphics[width=0.9\textwidth]{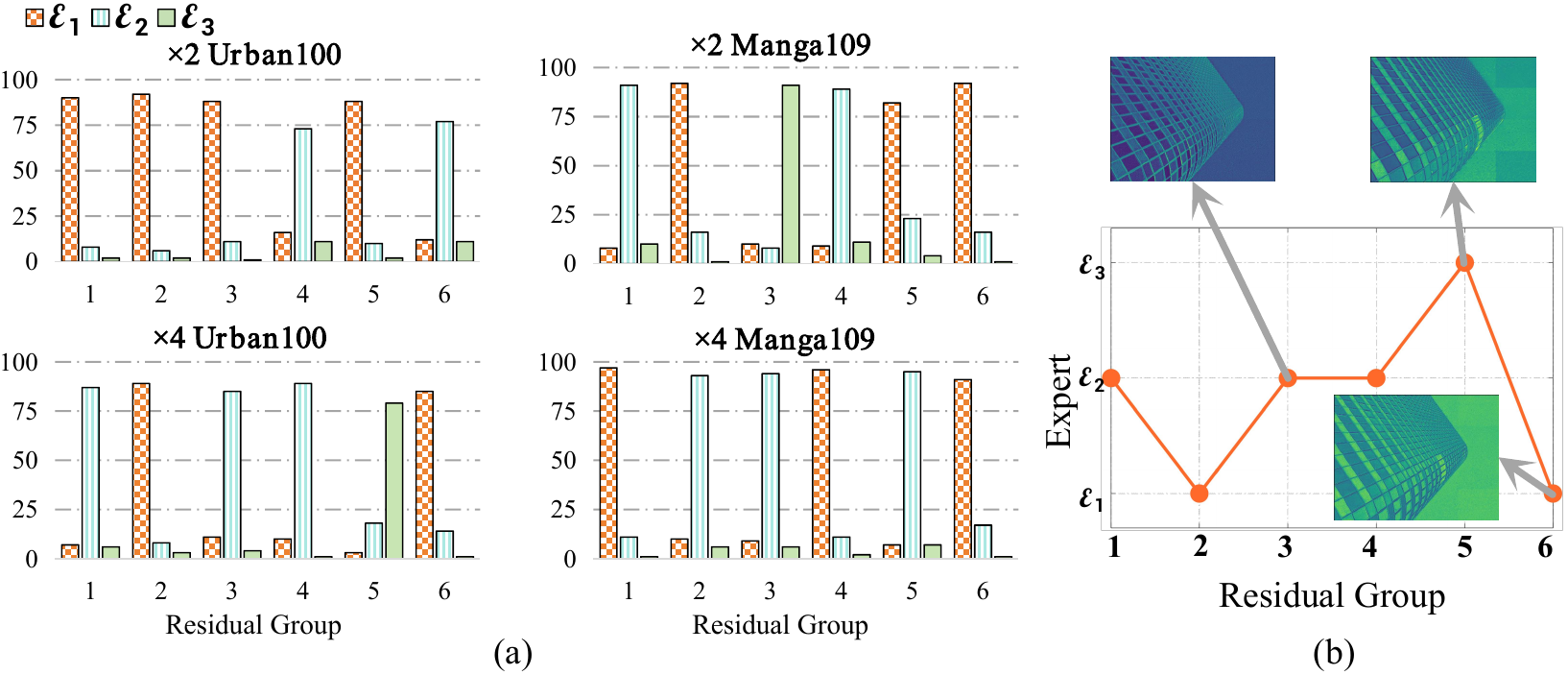}
  \caption{Analysis of adaptive expert routing in MSS-MoE. (a) Routing preferences of scale-specific experts across network depth, datasets, and SR scales. (b) Representative output features of experts operating at different region-level modeling granularities.}
  \label{fig:era}
  \vspace{-0.8cm}
\end{figure*}

\textbf{Superpixel Construction.} Table \ref{spab} compares different region construction strategies, including SLIC \cite{slic}, STA \cite{sta}, and the differentiable SSN \cite{ssn} adopted in our framework. Although STA achieves slightly higher reconstruction accuracy, it requires 460K parameters and 38G GMACs. In contrast, SSN achieves 27.69 dB on BSD100 with only 271K parameters and 22G GMACs, providing a more favorable balance between reconstruction quality and computational cost. Compared with SLIC, SSN also improves accuracy while requiring lower computation. We therefore adopt SSN as the default differentiable grouping mechanism.

\textbf{Region Budget.} Table \ref{tab:numsp} studies the influence of the number of superpixels used by MSS-MoE. The configuration [16,32,64] slightly reduces computation but decreases Urban100 PSNR to 26.28 dB, whereas increasing the budgets to [32,64,128] raises the cost to 39G GMACs without corresponding performance gains. Among the evaluated settings, [64,64,64] achieves the best performance-efficiency balance and is therefore adopted in the final model. Importantly, the number of regions is fixed only to maintain a predictable computational budget, while the spatial support and pixel membership of each region remain content dependent. Thus, a fixed region budget does not imply a fixed spatial partition.

\textbf{Feature Gating.} Fig. \ref{fig:spssm} evaluates different aggregation strategies within SP-SSM. Compared with removing the gate or replacing it with simple addition or concatenation, the proposed gated formulation achieves the best performance, reaching 32.22 dB on Urban100 and 39.01 dB on Manga109 for $\times2$ SR. This indicates that multiplicative gating provides a more effective mechanism for integrating the region-level global response with the dense feature branch.

\textbf{Local Mixed Attention.} As shown in Table \ref{tab:ab3}, combining SwinMHSA and CA in LMA achieves 27.69/26.40 dB on BSD100/Urban100, outperforming SwinMHSA alone (27.67/26.28 dB) and CA alone (27.65/26.22 dB). This confirms the complementarity of spatial and channel-wise local refinement.

\begin{table}[t]
\vspace{-0.5cm}
\centering
\caption{Analysis of the generalization capability of SP-MoMamba. We provide a performance comparison of SP-MoMamba on the low-light image enhancement task.}
\label{llie}
% \resizebox{\columnwidth}{!}{%

% \footnotesize
\setlength{\tabcolsep}{3pt}
\renewcommand{\arraystretch}{1.1}

\begin{tabular}{lcccc}
\hline
{\color[HTML]{1F1F1F} \textbf{Method}}                            & \multicolumn{2}{c}{{\color[HTML]{1F1F1F} \textbf{LOLv1}}}                       & \multicolumn{2}{c}{{\color[HTML]{1F1F1F} \textbf{LOLv2-Real}}}                  \\
{\color[HTML]{1F1F1F} \textbf{}}                     & {\color[HTML]{1F1F1F} \textbf{PSNR ↑}} & {\color[HTML]{1F1F1F} \textbf{SSIM ↑}} & {\color[HTML]{1F1F1F} \textbf{PSNR ↑}} & {\color[HTML]{1F1F1F} \textbf{SSIM ↑}} \\ \hline
{\color[HTML]{1F1F1F} UHDFour \cite{uhdfour}}             & {\color[HTML]{1F1F1F} 22.89}           & {\color[HTML]{1F1F1F} 0.8147}          & {\color[HTML]{1F1F1F} 19.42}           & {\color[HTML]{1F1F1F} 0.7896}          \\
{\color[HTML]{1F1F1F} Retinexformer \cite{retinexformer}}      & {\color[HTML]{1F1F1F} 22.71}           & {\color[HTML]{1F1F1F} 0.8177}          & {\color[HTML]{1F1F1F} 22.79}           & {\color[HTML]{1F1F1F} 0.8397}          \\
{\color[HTML]{1F1F1F} DMFourLLIE \cite{dmfourllie}}       & {\color[HTML]{1F1F1F} 22.98}           & {\color[HTML]{1F1F1F} 0.8273}          & {\color[HTML]{1F1F1F} 22.71}           & {\color[HTML]{1F1F1F} 0.8583}          \\
{\color[HTML]{1F1F1F} UHDFormer \cite{uhdformer}}         & {\color[HTML]{1F1F1F} 22.88}           & {\color[HTML]{1F1F1F} 0.8370}           & {\color[HTML]{1F1F1F} 19.71}           & {\color[HTML]{1F1F1F} 0.8320}           \\
{\color[HTML]{1F1F1F} RetinexMamba \cite{retinexmamba}} & {\color[HTML]{1F1F1F} 23.15}           & {\color[HTML]{1F1F1F} 0.8210}           & {\color[HTML]{1F1F1F} 21.73}           & {\color[HTML]{1F1F1F} 0.8290}           \\
{\color[HTML]{1F1F1F} MambaLLIE \cite{mamballie}}         & {\color[HTML]{1F1F1F} 22.80}            & {\color[HTML]{1F1F1F} 0.8315}          & {\color[HTML]{1F1F1F} 21.85}           & {\color[HTML]{1F1F1F} 0.8276}          \\
{\color[HTML]{1F1F1F} CWNet \cite{cwnet}}            & {\color[HTML]{1F1F1F} \underline{23.60}}            & {\color[HTML]{1F1F1F} 0.8496}          & {\color[HTML]{1F1F1F} 23.31}           & {\color[HTML]{1F1F1F} \underline{0.8641}}          \\
{\color[HTML]{1F1F1F} CIDNet \cite{cidnet}}             & {\color[HTML]{1F1F1F} \textbf{23.81}}           & {\color[HTML]{1F1F1F} \textbf{0.8574}} & {\color[HTML]{1F1F1F} \underline{23.43}}  & {\color[HTML]{1F1F1F} 0.8622}          \\
SP-MoMamba-T                                                & 23.25                                  & \underline{0.8570}                                  & \textbf{23.82}                                  & \textbf{0.8722}                                 \\ \hline
\end{tabular}%
% }
\vspace{-0.4cm}
\end{table}

\begin{table}[t]
\centering
\caption{Quantitative PSNR comparison with state-of-the-art methods for Gaussian color image denoising at $\sigma = 15$.}
\label{tab:denoising_results}

\setlength{\tabcolsep}{3pt}
\renewcommand{\arraystretch}{1.1}

\begin{tabular}{l|cccc}
\hline
Method & CBSD68 & Kodak24 & McMaster & Urban100 \\ \hline
FFDNet \cite{FFDNet} & 33.87 & 34.63 & 34.66 & 33.83 \\
DnCNN \cite{DnCNN} & 33.90 & 34.60 & 33.45 & 32.98 \\
DRUNet \cite{DRUNet} & 34.30 & 35.31 & 35.40 & 34.81 \\
SwinIR \cite{swinir} & 34.42 & 35.34 & 35.61 & 35.13 \\
Restormer \cite{Restormer} & 34.40 & 35.35 & 35.61 & 35.13 \\
Xformer \cite{Xformer} & 34.43 & 35.39 & 35.68 & 35.29 \\
MambaIR \cite{mambair} & {34.48} & {35.42} & {35.70} & {35.37} \\
\rowcolor[gray]{0.9} MambaIRv2 \cite{mambairv2} & {34.48} & {35.43} & {35.73} & {35.42} \\ \hline
SP-MoMamba & \textbf{34.51} & \textbf{{35.44}} & \textbf{{35.74}} & \textbf{{35.45}} \\ \hline
\end{tabular}
\vspace{-0.2cm}
\label{dn}
\end{table}

\begin{table*}[h]
\centering
\caption{Quantitative comparison for JPEG compression artifact reduction at different quality factors $q$.}
\label{tab:jpeg_results_updated}
\resizebox{\textwidth}{!}{
\begin{tabular}{c|c|cc|cc|cc|cc|cc|cc|cc} % 增加兩列 cc
\hline
\multirow{2}{*}{Dataset} & \multirow{2}{*}{$q$} & \multicolumn{2}{c|}{RNAN \cite{RNAN}} & \multicolumn{2}{c|}{RDN \cite{RDN}} & \multicolumn{2}{c|}{DRUNet \cite{DRUNet}} & \multicolumn{2}{c|}{SwinIR \cite{swinir}} & \multicolumn{2}{c|}{MambaIR \cite{mambair}} & \multicolumn{2}{c|}{MambaIRv2 \cite{mambairv2}} & \multicolumn{2}{c}{SP-MoMamba} \\
 & & PSNR & SSIM & PSNR & SSIM & PSNR & SSIM & PSNR & SSIM & PSNR & SSIM & PSNR & SSIM & PSNR & SSIM \\ \hline
\multirow{3}{*}{Classic5} & 10 & 29.96 & 0.8178 & 30.00 & 0.8188 & 30.16 & 0.8234 & 30.27 & 0.8249 & 30.27 & 0.8256 & 30.37 & 0.8269 & \textbf{30.41} & \textbf{0.8271} \\
 & 30 & 33.38 & 0.8924 & 33.43 & 0.8930 & 33.59 & 0.8949 & 33.73 & 0.8961 & 33.74 & 0.8965 & 33.81 & 0.8970 & \textbf{33.85} & \textbf{0.8972} \\
 & 40 & 34.27 & 0.9061 & 34.27 & 0.9061 & 34.41 & 0.9075 & 34.52 & 0.9082 & 34.53 & 0.9084 & 34.64 & 0.9093 & \textbf{34.68} & \textbf{0.9096} \\ \hline
\multirow{3}{*}{LIVE1} & 10 & 29.63 & 0.8239 & 29.67 & 0.8247 & 29.79 & 0.8278 & 29.86 & 0.8287 & 29.88 & 0.8301 & 29.91 & 0.8301 & \textbf{29.94} & \textbf{0.8303} \\
 & 30 & 33.45 & 0.9149 & 33.51 & 0.9153 & 33.59 & 0.9166 & 33.69 & 0.9174 & 33.72 & 0.9179 & 33.73 & 0.9179 & \textbf{33.77} & \textbf{0.9181} \\
 & 40 & 34.47 & 0.9299 & 34.51 & 0.9302 & 34.58 & 0.9312 & 34.67 & 0.9317 & 34.70 & 0.9320 & 34.73 & 0.9323 & \textbf{34.78} & \textbf{0.9325} \\ \hline
\end{tabular}
}
\label{car}
\end{table*}

\begin{figure*}[t]
    \centering
    \includegraphics[width=\textwidth]{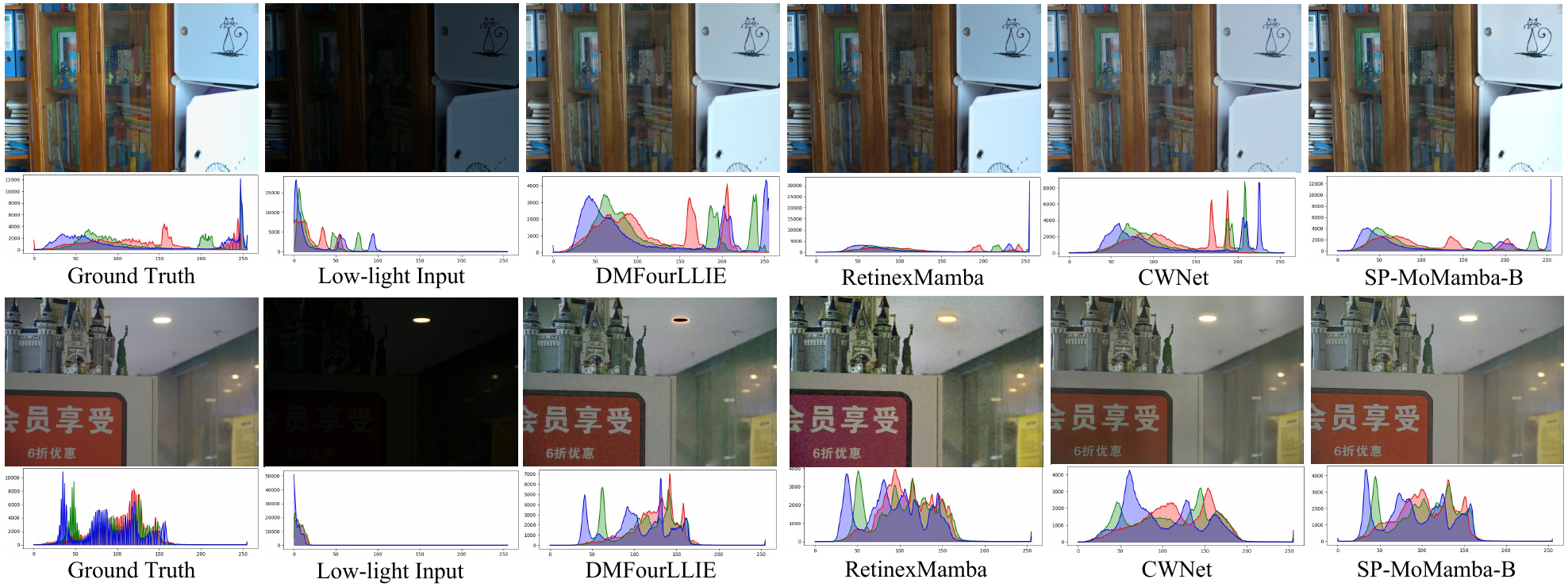}
    % \vspace{-0.2cm}
    \caption{Qualitative comparisons with existing LLIE methods on the LOLv1 and LOLv2-Real benchmark datasets \cite{RetinexNet}. }
    \label{fig:llie}
    \vspace{-0.5cm}
\end{figure*}

\subsection{Analysis of Adaptive Expert Routing}
To further understand the behavior of MSS-MoE, Fig. \ref{fig:era}(a) analyzes expert routing decisions across network depth, datasets, and upscaling factors. The router exhibits non-uniform expert preferences: different residual groups activate different scale-specific experts, and the routing distributions vary between Urban100 and Manga109 as well as between $\times2$ and $\times4$ SR. This behavior is consistent with the intended role of MSS-MoE. Since each expert corresponds to a different region-level modeling granularity, varying routing decisions indicate that a single fixed representation scale is not uniformly preferred across all features and restoration settings. Coarser-scale experts are more frequently selected in some stages, whereas finer-scale experts dominate in others, suggesting that the required abstraction level changes throughout feature reconstruction. Fig. \ref{fig:era}(b) further visualizes intermediate outputs from different experts. The resulting feature responses exhibit different structural characteristics across scales and network depths, providing qualitative evidence that the experts do not collapse to identical representations. Together with the scale ablation in Fig. \ref{fig:abstudy}(c), these routing analyses support the motivation of MSS-MoE: dynamically selecting the region-level modeling granularity is more flexible than imposing a single fixed scale throughout the network.

\subsection{Generalization on Other Restoration Tasks}
% To evaluate the generalization capability of our method, we extend its application to the low-light image enhancement (LLIE) task, as shown in Table \ref{llie}. We compare SP-MoMamba-T against comprehensive state-of-the-art approaches, including Transformer-based (e.g., Retinexformer) and CNN-based (e.g., CIDNet) models on the LOLv1 and LOLv2-Real datasets \cite{RetinexNet}. Our model demonstrates clear improvements, particularly on the more challenging LOLv2-Real benchmark. Specifically, SP-MoMamba-T outperforms the competitive CIDNet method on LOLv2-Real, achieving improvements of 0.39 dB in PSNR and 0.010 in SSIM. We attribute this to the strong global modeling capability of our superpixel-based architecture, which effectively restores illumination while preserving structural details. Furthermore, the visual comparisons in Fig. \ref{fig:llie} confirm that our method successfully recovers visibility and accurate color distributions in real-world dark scenarios. Overall, these results highlight the flexibility and strong generalization potential of our design beyond standard super-resolution tasks.
To examine whether the proposed region-level state-space modeling framework generalizes beyond super-resolution, we further evaluate SP-MoMamba on low-light image enhancement (LLIE), Gaussian color image denoising, and JPEG compression artifact reduction.

\textbf{Low-Light Image Enhancement. }Table \ref{llie} compares SP-MoMamba-T with representative CNN-based methods (e.g., UHDFour \cite{uhdfour}, DMFourLLIE \cite{dmfourllie}, CIDNet \cite{cidnet}), Transformer-based methods (e.g., Retinexformer \cite{retinexformer}, UHDFormer \cite{uhdformer}) and Mamba-based methods (e.g., RetinexMamba \cite{retinexmamba}, MambaLLIE \cite{mamballie}, CWNet \cite{cwnet}) on LOLv1 and LOLv2-Real \cite{RetinexNet}. SP-MoMamba-T achieves 23.82 dB/0.8722 on LOLv2-Real, outperforming the compared methods under both PSNR and SSIM. Fig. \ref{fig:llie} further shows that the model restores visibility and structural details under challenging real-world illumination conditions. These results suggest that compact region-level interaction remains effective for spatially non-uniform restoration.

% As shown in Table \ref{llie}, we compare SP-MoMamba-T against comprehensive state-of-the-art approaches, including Transformer-based (e.g., Retinexformer \cite{retinexformer}, UHDFormer \cite{uhdformer}) and CNN-based (e.g., UHDFour \cite{uhdfour}, DMFourLLIE \cite{DMNet}, CIDNet \cite{cidnet}) and Mamba-based (e.g. RetinexMamba \cite{retinexmamba}, MambaLLIE \cite{mamballie}, CWNet \cite{cwnet}) models on the LOLv1 and LOLv2-Real datasets \cite{RetinexNet}. Our model demonstrates clear improvements, particularly on the more challenging LOLv2-Real benchmark. Specifically, SP-MoMamba-T outperforms the competitive CIDNet method on LOLv2-Real, achieving improvements of 0.39 dB in PSNR and 0.010 in SSIM. We attribute this to the strong global modeling capability of our superpixel-based architecture, which effectively restores illumination while preserving structural details. Furthermore, the visual comparisons in Fig. \ref{fig:llie} confirm that our method successfully recovers visibility and accurate color distributions in real-world dark scenarios.

\textbf{Gaussian Color Image Denoising.} Table \ref{dn} reports results for Gaussian color denoising with $\sigma=15$, including CNN-based methods (FFDNet \cite{FFDNet}, DnCNN \cite{DnCNN}, DRUNet \cite{DRUNet}), Transformer-based methods (SwinIR \cite{swinir}, Restormer \cite{Restormer}, Xformer \cite{Xformer}), and Mamba-based methods (MambaIR \cite{mambair}, MambaIRv2 \cite{mambairv2}). SP-MoMamba achieves the highest PSNR among the compared methods on CBSD68 \cite{B100}, Kodak24 \cite{Kodak}, McMaster \cite{McMaster}, and Urban100 \cite{Urban100}, with improvements of 0.03, 0.01, 0.01, and 0.03 dB over MambaIRv2, respectively. Rather than attributing these gains to “semantic scanning,” the results indicate that content-adaptive region-level interaction can be transferred to restoration tasks that require simultaneously suppressing degradation and preserving spatial structures.

% To further validate the versatility of SP-MoMamba, we conduct experiments on the Gaussian color image denoising task at noise level $\sigma = 15$, as reported in Table \ref{dn}. We compare our method against a broad set of representative baselines, including CNN-based methods (FFDNet \cite{FFDNet}, DnCNN \cite{DnCNN}, DRUNet \cite{DRUNet}), Transformer-based methods (SwinIR \cite{swinir}, Restormer \cite{Restormer}, Xformer \cite{Xformer}), and Mamba-based methods (MambaIR \cite{mambair}, MambaIRv2 \cite{mambairv2}), evaluated on four standard benchmarks: CBSD68, Kodak24, McMaster, and Urban100. SP-MoMamba consistently achieves the highest PSNR scores across all four datasets. Notably, SP-MoMamba surpasses MambaIRv2 by 0.03 dB on CBSD68, 0.01 dB on Kodak24, 0.01 dB on McMaster, and 0.03 dB on Urban100, demonstrating that the proposed superpixel-driven semantic scanning mechanism benefits not only super-resolution tasks but also noise suppression tasks that require precise preservation of structural boundaries and texture fidelity. The consistent improvement across all benchmarks reflects that grouping semantically coherent pixels into superpixel tokens effectively guides the model to distinguish between genuine image structures and noise patterns during restoration.

\textbf{JPEG Compression Artifact Reduction.} 
We further evaluate SP-MoMamba on the task of JPEG compression artifact reduction under multiple quality factors ($q = 10, 30, 40$) on the Classic5 \cite{Classic} and LIVE1 \cite{LIVE1} benchmark datasets, as presented in Table \ref{car}. We compare against RNAN \cite{RNAN}, RDN \cite{RDN}, DRUNet \cite{DRUNet}, SwinIR \cite{swinir}, MambaIR \cite{mambair}, and MambaIRv2 \cite{mambairv2}. SP-MoMamba consistently obtains the highest PSNR/SSIM values among the compared approaches across the evaluated settings. The improvement under low-quality compression indicates that region-level modeling remains effective when the input contains strong structured artifacts and local discontinuities.

Overall, the consistent results across SR, LLIE, denoising, and JPEG artifact reduction indicate that the proposed content-aware region-level state-space modeling framework is not restricted to a specific SR degradation and exhibits promising generalization across image restoration tasks.

\section{Conclusion}
% In this work, we propose SP-MoMamba, an efficient image restoration framework based on content-aware region-level state-space modeling. Rather than solely redesigning traversal paths over dense image features, SP-SSM constructs compact region-level representations from spatially coherent features and performs global sequence modeling over these content-aware regions, reducing redundant computation while facilitating long-range structural interaction. MSS-MoE further addresses the granularity mismatch of region-level modeling by dynamically selecting scale-specific SP-SSM experts, enabling adaptive multi-granularity global modeling. LSME complements this abstraction with dense local refinement for fine-detail reconstruction. Extensive experiments demonstrate that SP-MoMamba achieves strong reconstruction performance with a favorable balance among model size, computational cost, inference latency, and memory consumption, while its consistent performance across multiple restoration tasks further indicates the generalization potential of the proposed region-level modeling framework.
In this work, we propose SP-MoMamba, an efficient image restoration framework based on content-aware region-level state-space modeling. Rather than solely redesigning traversal paths over dense image features, SP-SSM constructs compact region-level representations from spatially coherent features and performs global sequence modeling over these content-aware regions, reducing redundant computation while facilitating long-range structural interaction. MSS-MoE further addresses the granularity mismatch of region-level modeling by dynamically selecting scale-specific SP-SSM experts, enabling adaptive multi-granularity global modeling. LSME complements this abstraction with dense local refinement for fine-detail reconstruction. Extensive experiments demonstrate that SP-MoMamba achieves strong reconstruction performance with a favorable balance among model size, computational cost, inference latency, and memory consumption, while its consistent performance across multiple restoration tasks further indicates the generalization potential of the proposed region-level modeling framework. 

In addition, our results suggest that visual sequence construction and representation granularity are important design factors for efficient SSM-based restoration. Combining adaptive region-level interaction with dense local refinement reduces redundant global computation while retaining spatial detail. The ablation results further indicate that selecting suitable modeling granularities is more effective than simply increasing expert capacity or activating additional experts. Nevertheless, overly coarse region representations can compromise fine-detail reconstruction, while the predefined scale set and fixed superpixel count constrain representation flexibility. Future work will explore content-adaptive region allocation and scale selection to better balance compactness and detail preservation, alongside sparse training and hardware-aware implementations to improve practical efficiency.

\section{Data availability statement}
The 15 datasets used in this paper are publically available (\textit{i.e.} DIV2K \cite{DIV2K}, Flickr2K \cite{EDSR}, Set5 \cite{Set5}, Set14 \cite{Set14}; BSD100 \cite{B100}, Urban100 \cite{Urban100}, Manga109 \cite{manga109}, RealSRv3 \cite{realsr}, LOLv1 \cite{RetinexNet}, LOLv2-Real \cite{RetinexNet}, CBSD68 \cite{CBSD68}, Kodak24 \cite{Kodak}, McMaster \cite{McMaster}, Classic5 \cite{Classic}, and LIVE1 \cite{LIVE1}). The datasets analyzed during this study are all publicly available for research purposes.

\section*{Acknowledgements}
This work was completed by Wenbin Zou under the joint guidance of Huiping Zhuang and Yi Wang. This research was supported by the National Natural Science Foundation of China (62306117), the Guangdong Basic and Applied Basic Research Foundation (2026A1515011478), and GJYC program of Guangzhou (2024D03J0005). The research work described in this paper was conducted in the JC STEM Lab of Machine Learning and Computer Vision funded by the Hong Kong Jockey Club Charities Trust. This research received partial support from the Global STEM Professorship Scheme from the Hong Kong Special Administrative Region.

\bibliography{reference}
%\bibliographystyle{plain}
%\bibliography{reference}
\newpage
\appendixpage
\section{Pseudocode and Pipeline of the MSS-MoE Mechanism}
Algorithm 1 provides a detailed demonstration of the forward propagation process of the Multi-Scale Superpixel Mixture of State Space Experts (MSS-MoE). This algorithm receives two input features, $x_1$ and $x_2$. First, it dynamically calculates the weight of each expert using a routing network followed by a Softmax function. Then, it uses the top-$k$ mechanism to select core experts and reset the remaining weights to zero to achieve sparse activation. During the training phase, the algorithm evaluates all experts and feeds the feature $x_1$ along with the unique scale factor $s_i$ of each expert into the superpixel state space module (SP-SSM) for region-level feature modeling. It then multiplies the region feature $\text{SP-SSM}(x_1, s_i)$ element-wise by the gating feature $\sigma (x_2)$, and obtains the output through weighted summation. In the inference stage, the algorithm evaluates only the selected top-k experts and directly calculates and outputs the weighted results of the selected top-$k$ experts. This dual-branch design enables the model to maintain good feature flow during training and significantly reduce computational cost during inference, achieving efficient multi-scale long-range context modeling.

\begin{algorithm}[t]
\caption{Multi-Scale Superpixel Mixture of State Space Experts (MSS-MoE)}
\begin{algorithmic}[1]
\STATE \textbf{Input:} Features $x_1$ and $x_2$
\STATE \textbf{Parameters:} $n$ experts $\mathcal{E} = \{\mathcal{E}_i\}_{i=1}^n$, router $\mathcal{G}$, scale set $\mathcal{S} = \{s_i\}_{i=1}^n$, Superpixel state-space module (SP-SSM), and the number $k$ of activated experts.
% \STATE $k = 1$
\STATE Compute router outputs: $g = \mathcal{G}(x_2)$
\STATE Normalize weights: $w = \text{Softmax}(g)$
\STATE Select top-$k$ experts: $w_{\text{top-}k}, \mathcal{I}_k = \text{topk}(w, k)$
\STATE Set all other weights to zero: $w_i = 0$, $\forall i \notin \mathcal{I}_k$
\IF {training} 
    \FOR {\textbf{each} $i \in \mathcal{E}$}
        \STATE $y_i = \text{SP-SSM}(x_1, s_i)\odot \sigma(x_2)$
    \ENDFOR
    \STATE Compute final output: $y = \sum_{i=1}^n w_i \cdot y_i$
\ELSE
    \STATE Compute final output: $y=\sum_{i\in \mathcal{I}_k} w_iy_i$.
\ENDIF
\STATE \textbf{Output:} Final output $y$
\end{algorithmic}
\label{alg:mss-moe}
\end{algorithm}

\section{Further Implementation Details}
Table \ref{tab:imp} provides a detailed list of the specific implementation details and training hyperparameter configurations for two model versions, SP-MoMamba-T and SP-MoMamba-B. In terms of model architecture, the two are distinguished in the core dimension, where SP-MoMamba-T contains 3 RGs with a channel dimension of 36, whereas SP-MoMamba-B contains 4 RGs with a channel dimension of 48. Both variants use two SGMEs and two LSMEs per RG, an MLP ratio of 2, 64 superpixels, and an MoE containing three experts with top-$1$ routing over the scale set \([1, 2, 4]\). In terms of training settings, the two models are completely identical: both are based on the DIV2K and Flickr2K combined datasets and trained using the Adam optimizer for 500K iterations, with a batch size of 32 and an initial learning rate of 1.00e-03. They undergo a stepwise decay at 250K, 400K, 450K, and 475K iterations with a decay rate of 0.5, while introducing an FFT (Fast Fourier Transform) loss function with a weight of 0.1 to constrain the quality of high-frequency reconstruction. Our implementation leverages the publicly available PyTorch-based BasicSR\footnote{\url{https://github.com/XPixelGroup/BasicSR}} framework for both architecture design and training. Additionally, we employ the fvcore\footnote{\url{https://github.com/facebookresearch/fvcore}} Python package to calculate GMACs and parameter counts.

\begin{table}[ht]
\centering
\caption{Implementation Details.}

\footnotesize
\setlength{\tabcolsep}{2pt}
\renewcommand{\arraystretch}{1.08}
% \resizebox{0.6\columnwidth}{!}{%
\begin{tabular}{@{}lcc@{}}
\toprule
Parameters            & \multicolumn{1}{l}{SP-MoMamba-T} & \multicolumn{1}{l}{SP-MoMamba-B} \\ \midrule
Num. RGs              & 3                                & 4                                \\
Num. SGMEs and LSMEs  & 2                                & 2                                \\
Channel dimension     & 36                               & 48                               \\
MLP ratio            & \multicolumn{2}{c}{2}                                               \\
Num. superpixels      & \multicolumn{2}{c}{64} \\
Number of experts $\mathcal{E}$        & \multicolumn{2}{c}{3}                                               \\
Number of activated experts $k$         & \multicolumn{2}{c}{1}                                               \\
Scale list            & \multicolumn{2}{c}{{[}1, 2, 4{]}}                                     \\
Training datasets      & \multicolumn{2}{c}{DIV2K+Flickr2K}                                  \\
Optimizer             & \multicolumn{2}{c}{Adam}                                            \\
Batch size            & \multicolumn{2}{c}{32}                                              \\
Total iterations & \multicolumn{2}{c}{500K}                                            \\
FFT loss weight      & \multicolumn{2}{c}{0.1}                                             \\
Learning Rate         & \multicolumn{2}{c}{1.00E-03}                                        \\
LR Decay Rate         & \multicolumn{2}{c}{0.5}                                             \\
LR Decay Milestones   & \multicolumn{2}{c}{{[}250K, 400K, 450K, 475K{]}}                     \\ \bottomrule
\end{tabular}%
% }
\label{tab:imp}
\vspace{-0.6cm}
\end{table}

\begin{table}[ht]
\centering
\caption{Ablation study of loss function. We report PSNR (dB $\uparrow$) on the Y channel for ×4 upscaling.}
\label{tab:loss}
% \resizebox{0.5\columnwidth}{!}{%
\begin{tabular}{@{}ccccc@{}}
\toprule
\multicolumn{1}{l}{Method}    & $L_1$  & FFT & BSD100 & Urban100 \\ \midrule
\multirow{4}{*}{SP-MoMamba-T} & 1.0 & 0.0 & 27.65  & 26.26    \\
                              & \textbf{1.0} & \textbf{0.1} & \textbf{27.69}  & \textbf{26.40}    \\
                              & 1.0 & 0.2 & 27.68  & 26.32    \\ 
                              & 1.0 & 0.3 & 27.61  & 26.14  \\ \bottomrule
\end{tabular}%
\vspace{-0.6cm}
% }
\end{table}

\begin{figure*}[ht]
    \centering
    \includegraphics[width=0.85\textwidth]{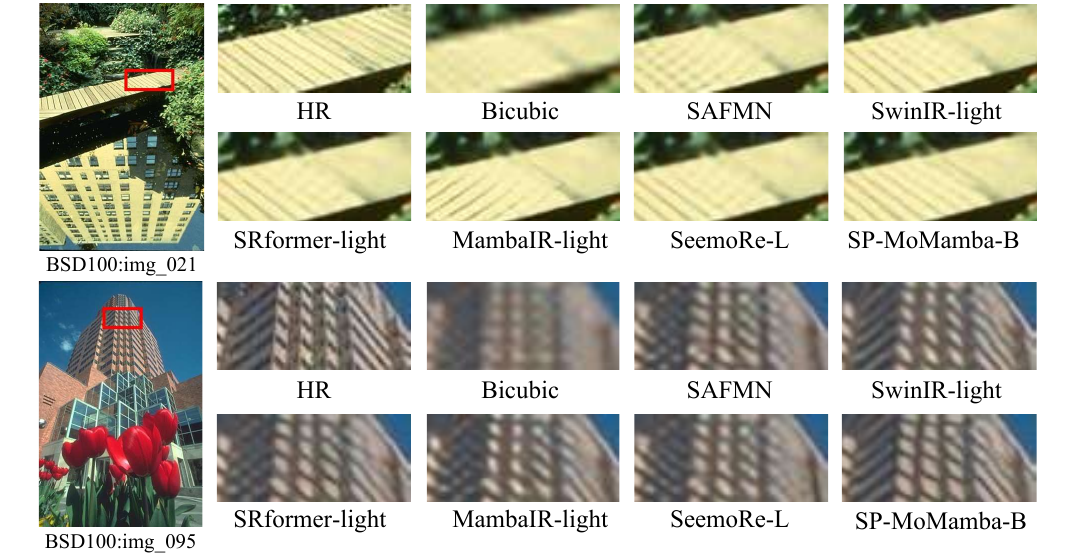}
    \caption{Visual comparison of SP-MoMamba-B with state-of-the-art methods on challenging cases for ×4 SR from the BSD100 benchmark.}
    \label{fig:supp1}
\end{figure*}

\begin{figure*}[ht]
    \centering
    \includegraphics[width=0.85\textwidth]{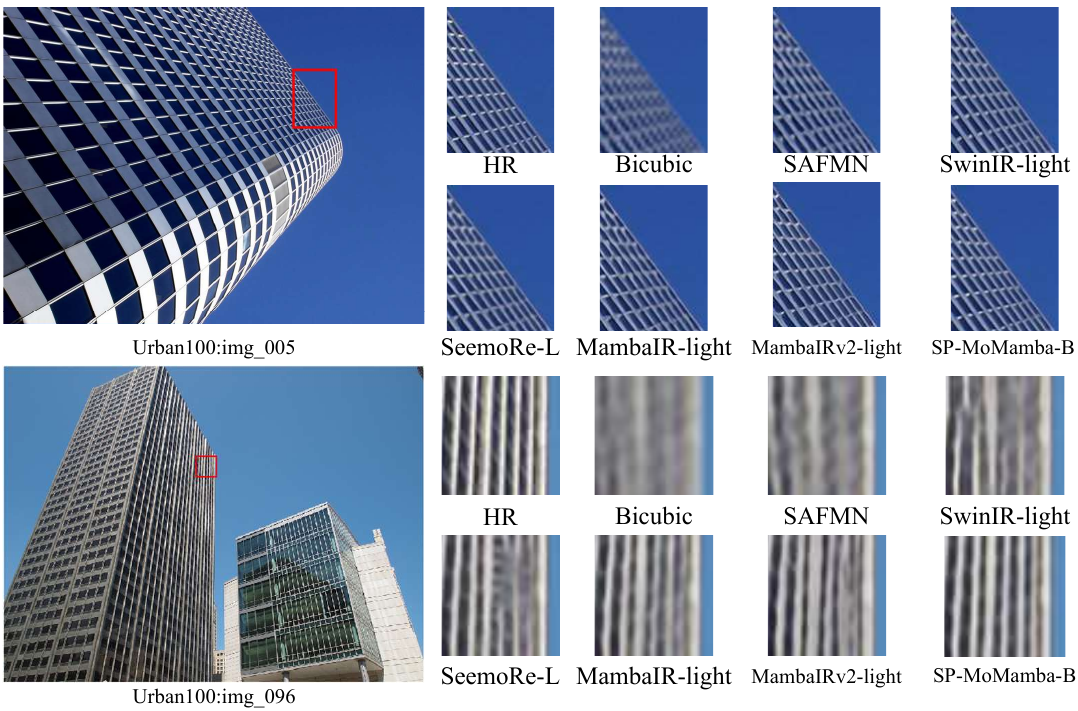}
    \caption{Visual comparison of SP-MoMamba-B with state-of-the-art methods on challenging cases for ×4 SR from the Urban100 benchmark.}
    \label{fig:supp2}
\end{figure*}

\section{Loss Function and Ablation Study}
We use FFT loss and $L_1$ loss to jointly optimize the network. The specific expression is as follows:
\begin{align}
    L_{total} &= L_1 + w*L_{FFT}\\
    L_1 &= ||I_{gt}-I_{SR}||_1\\
    L_{FFT} &= ||FFT(I_{gt})-FFT(I_{SR})||_1 
\end{align}
where $FFT(\cdot)$ denotes the Fourier transform. The $I_{gt}$ and $I_{SR}$ represent the ground-truth image and super-resolution image. To verify the effectiveness of the loss function, we conducted a set of ablation experiments, as shown in Table \ref{tab:loss}. As can be seen from the table, compared with using only $L_1$ loss, incorporating the FFT loss imposes additional frequency-domain constraints, thereby alleviating texture oversmoothing and improving reconstruction performance.

\section{More Visual Results}
Figures \ref{fig:supp1} and \ref{fig:supp2} show the visual comparison results of SP-MoMamba-B with multiple state-of-the-art (SOTA) super-resolution models on the BSD100 and Urban100 benchmark datasets for $\times 4$ SR. When dealing with highly challenging high-frequency scenes containing dense parallel lines, complex repeated grids, and strong perspective deformations, competing methods such as Bicubic, SAFMN \cite{SAFMN}, SwinIR-Light \cite{swinir}, and MambaIR variants \cite{mambair, mambairv2} all exhibit varying degrees of serious artifacts such as detail blurring, line breaks, aliasing, or moiré patterns. In contrast, SP-MoMamba-B better preserves geometric structures and reconstructs clearer, straighter, and more regularly arranged texture patterns. It effectively suppresses artifacts while producing edge sharpness and overall visual quality closer to the ground-truth HR images.

\end{document}